\documentclass[lettersize,journal]{IEEEtran}
\usepackage{multirow}%
\usepackage{amsmath,amsfonts}
\usepackage{algorithmic}
\usepackage{algorithm}
\usepackage{array}
\usepackage[caption=false,font=normalsize,labelfont=sf,textfont=sf]{subfig}
\usepackage{textcomp}
\usepackage{stfloats}
\usepackage{url}
\usepackage{verbatim}
\usepackage{graphicx}
\usepackage{cite}
\usepackage{amssymb}
\usepackage{pifont}
\newcommand{\xmark}{\ding{55}}%
\usepackage{xcolor}

\DeclareMathOperator*{\argmax}{\arg\!\max}

\usepackage{caption}
\usepackage{subcaption}
\usepackage{subfig}
\usepackage{stfloats}

\begin{document}

\title{AdaSemSeg: An Adaptive Few-shot Semantic Segmentation of Seismic Facies}

\author{Surojit~Saha\footnotemark
        ~and~Ross~Whitaker
\thanks{S. Saha, and R. Whitaker are with the Scientific Computing and Imaging Institute, Kahlert School of Computing, The University of Utah, Salt Lake City, USA. (e-mail:surojit.saha@utah.edu; rosstwhitaker@gmail.com)}
}



\maketitle

\begin{abstract}
Automated interpretation of seismic images using deep learning methods is challenging because of the limited availability of training data. Few-shot learning is a suitable learning paradigm in such scenarios due to its ability to adapt to a new task with limited supervision (small training budget). Existing few-shot semantic segmentation (FSSS) methods fix the number of target classes. Therefore, they do not support joint training on multiple datasets varying in the number of classes. In the context of the interpretation of seismic facies, fixing the number of target classes inhibits the generalization capability of a model trained on one facies dataset to another, which is likely to have a different number of facies. To address this shortcoming, we propose a few-shot semantic segmentation method for interpreting seismic facies that can adapt to the varying number of facies across the dataset, dubbed the \emph{AdaSemSeg}. In general, the backbone network of FSSS methods is initialized with the statistics learned from the ImageNet dataset for better performance. The lack of such a huge annotated dataset for seismic images motivates using a self-supervised algorithm on seismic datasets to initialize the backbone network. We have trained the AdaSemSeg on three public seismic facies datasets with different numbers of facies and evaluated the proposed method on multiple metrics. The performance of the AdaSemSeg on \emph{unseen} datasets (not used in training) is better than the prototype-based few-shot method and baselines.
\end{abstract}


\section{Introduction}
\IEEEPARstart{T}{he} study of facies in seismic images has emerged as an important topic of research in the recent past due to the comprehensive characterization of the earth's subsurface. Seismic facies represent regions in the earth's crust with similar geological characteristics, as indicated by correlated reflection properties in seismic images. In addition to identifying hydrocarbon reservoirs, delineating horizons in seismic images into facies finds potential applications in carbon capture and storage \cite{CS_Facies_Seg_1,CS_Facies_Seg_2}. Manual interpretation of seismic facies by expert geologists is a painstacking process. Moreover, the presence of complex morphological variations often leads to subjective interpretation. Thus, automated detection is encouraged for consistent and efficient solutions.

\begin{figure}[th]
\centering
\includegraphics[width=0.49\textwidth]{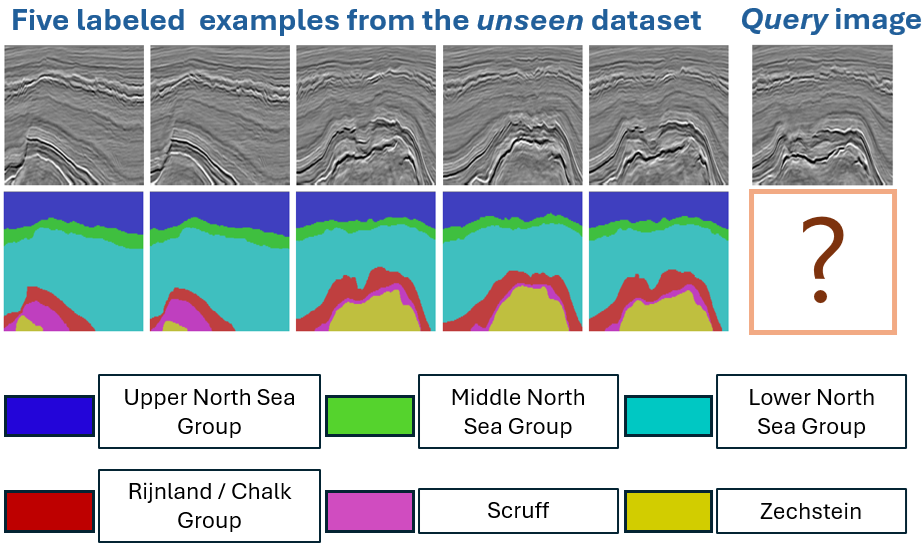}
\caption{In this figure, we illustrate the task of predicting the \emph{seismic facies} (a multi-class segmentation mask) in a \emph{query image} from an \emph{unseen} seismic dataset \cite{f3_facies_netherlands_2019} by a semantic segmentation model using \emph{a few annotated examples} from the dataset. This presents a realistic scenario for the interpretation of seismic facies. We address this problem using the \emph{few-shot semantic segmentation} method. 
} \label{fig:mot_FSSS_F3}
\end{figure}

Advancements in deep learning methods have accelerated the automated interpretation of seismic images \cite{f3_facies_netherlands_2019,seismic_facies_2022}. Supervised deep learning methods produce impressive results with a sufficiently large training dataset. However, obtaining a large annotated seismic dataset is costly and impractical in most scenarios \cite{FSSS_Facies,SAM_Salt_Bodies_1,SAM_Salt_Bodies_2,Facies_Self-Supervised}. Therefore, a realistic way to handle this situation is to interpret seismic images using a few labeled examples from a \emph{novel/unseen} seismic dataset. In this paper, we study the problem of interpreting \emph{seismic facies} using a few annotated examples from the unseen seismic dataset. In Fig. \ref{fig:mot_FSSS_F3}, we present the problem of interpreting seismic facies in the F3 facies dataset\cite{f3_facies_netherlands_2019} using a few annotated examples of the dataset. Predicting seismic facies in a \emph{query/test image} from the unseen dataset is a \emph{multi-class} segmentation problem. \emph{Transfer learning} (TL) \cite{Zeiler_Fergus,TL_CVPRW_2014} is a naive approach to solving the segmentation task with a few labeled examples. TL fine-tunes the semantic segmentation model parameters with a few examples from the new dataset. However, we always run the risk of overfitting the model parameters to the limited training data. Despite different regularization strategies, such as early-stopping, learning rate decay, weight decay, and dropout 
, the model fails to generalize to query/test examples. Therefore, we should consider methods that avoid fine-tuning parameters with the few annotated samples from the unseen target dataset.

\begin{figure}[th]
\centering
\includegraphics[width=0.49\textwidth]{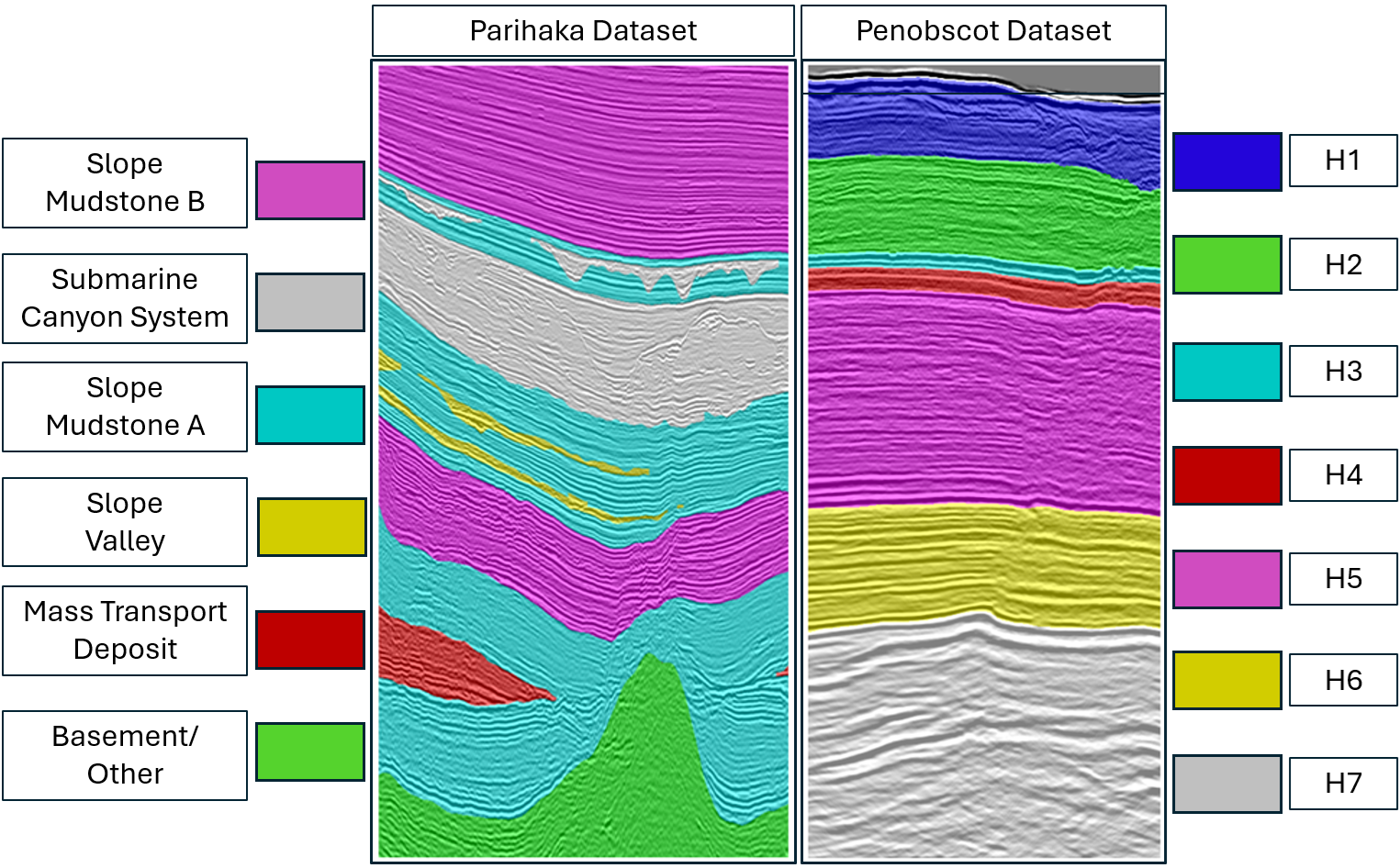}
\caption{This figure illustrates the heterogeneity in the number and types of facies across seismic datasets. In this example, we show the classification of the horizons of the Parihaka \cite{Parihaka_facies_new_zealand_2020} and Penobscot \cite{Penobscot_facies_canada_2021} datasets, where we observe a difference in the number of seismic facies and variation in the naming convention of the facies. 
} \label{fig:mot_Adaptive_FSSS}
\end{figure}

A potential approach to deal with the realistic scenario of limited training data would be the induction of a learning paradigm that can quickly adapt to the target task while having limited supervision. This learning technique is known as \emph{few-shot} learning \cite{Model_FSL_Laro_2017,Proto_FSL_Snell_2017,MAML_FSL_Finn_2017}. Few-shot learning involves two stages: the \emph{meta-training} and \emph{meta-testing}. In the meta-training stage, large annotated data, known as the \emph{source} data 
 is used to train a deep neural network. The meta-testing stage involves adapting the model to the new task \emph{without} fine-tuning the meta-trained parameters on the \emph{target} data. Instead, a few annotated samples (e.g., 1 or 5 samples) from the target dataset are used to guide the predictions on the remaining samples in the target dataset. 
Broadly, the few-shot learning algorithms are categorized as model-based (black-box) \cite{FSL_2017}, metric-based (non-parametric) \cite{Proto_2017} or optimization-based methods \cite{MAML_2017}. An overview of few-shot learning in the context of identifying seismic facies is presented in Section \ref{sec:FSSS}.

Predicting facies in a seismic dataset is a multi-class segmentation problem. In the context of few-shot learning, \emph{few-shot semantic segmentation} (FSSS) \cite{FSSS_ICIP_2022,FSSS_CVPRW_2023} methods do multi-class classification at the pixel level, an extension to the binary segmentation task \cite{shaban2017,dgpnet_2022}. Generally, the FSSS methods need to know \emph{a priori} the number of classes present in the dataset. For example, the number of seismic facies
in the Parihaka dataset \cite{Parihaka_facies_new_zealand_2020} is six, and the number of facies in the Penobscot dataset \cite{Penobscot_facies_canada_2021} is seven. Therefore, we must build separate FSSS models for the Parihaka and Penobscot datasets. This is a limitation of the existing FSSS methods that inhibits the training on multiple datasets varying in the number of target classes. We illustrate this problem in Fig. \ref{fig:mot_Adaptive_FSSS}, where we show the classification of the horizons of the Parihaka and Penobscot datasets based on the seismic data characteristics, such as reflection patterns and amplitude.

The community has recognized the limitation of the existing FSSS methods to deal with multiple datasets varying in the number of target classes, and new methods have been built in the recent past to alleviate this issue \cite{MSeg_2020,LMSEG_2023,DaTaSeg_2023}. The MSeg \cite{MSeg_2020} and LMSEG \cite{LMSEG_2023} build a unified taxonomy of classes present in multiple semantic segmentation datasets by interpreting the names of the class labels.
The viability of constructing a unified taxonomy for interpreting seismic facies is questionable due to the variability in the geological features observed across datasets. In addition to the difference in the number of seismic facies between the Parihaka and Penobscot datasets in Fig. \ref{fig:mot_Adaptive_FSSS}, we observe a variation in the naming convention of the facies, possibly indicating variation in the facies composition. 
All these limitations motivated us to devise a generalized few-shot segmentation method that is independent of the facies' names and can accommodate the variability in the number of facies across multiple datasets.

We propose an FSSS method for identifying seismic facies using Gaussian processes that can adapt to different numbers of facies across datasets, named the \emph{AdaSemSeg}. The proposed technique is motivated and adapted from the method introduced by Johnander \textit{et al.} \cite{dgpnet_2022} that empirically demonstrated the strength of the Gaussian process regression in the few-shot setup. In this work, we devised a technique that extends the capability of the method in \cite{dgpnet_2022} (designed for binary segmentation) to do multi-class segmentation (aka semantic segmentation). In the AdaSemSeg, the multi-class segmentation problem is divided into multiple binary segmentation tasks, where each task recognizes a particular type of facies in a given dataset using a shared backbone network. The number of binary tasks for a dataset depends on the number of facies present. All binary tasks use the same backbone network, i.e., the number of trainable parameters is fixed and does not vary with the number of binary tasks (representing facies). Using a shared backbone network for all binary tasks offers flexibility to the proposed AdaSemSeg to adapt to the varying number of facies. The final multi-class prediction for a query image is obtained by aggregating the outcomes of multiple binary segmentation tasks. 

In this work, we demonstrate the efficacy of the AdaSemSeg on three benchmark 3D facies datasets: the F3  from the Netherlands \cite{f3_facies_netherlands_2019}, Penobscot from Canada \cite{Penobscot_facies_canada_2021} and Parihaka from New Zealand \cite{Parihaka_facies_new_zealand_2020} having six, seven and six different facies, respectively. In our experiments, one of the datasets is considered the target data, and we train the AdaSemSeg on the remaining two datasets, which we call the source data. For all experimental results in this paper, the AdaSemSeg is evaluated on the target data using the statistics learned from the source data, which is assumed to be related but different from the target. To put things more clearly, the parameters of the AdaSemSeg are trained \emph{only} using samples in the source data, and we \emph{do not fine-tune} the AdaSemSeg on any samples in the target data. This makes the AdaSemSeg different from \emph{transfer learning} that fine-tunes the model parameters on samples from the target data. We compare the performance of the AdaSemSeg with different baselines using multiple evaluation metrics studied in the literature \cite{f3_facies_netherlands_2019,seismic_facies_2022}. In addition, the performance of the AdaSemSeg is compared with a prototype-based few-shot segmentation method developed for seismic facies \cite{facies_protoseg_2023} and a regular semantic segmentation network \emph{fine-tuned} on a few samples from the target dataset (aka transfer learning). A summary of the contributions is as follows:
\begin{itemize}
    \item We propose an adaptive FSSS method for identifying seismic facies that is flexible to handle the variability in the number of facies across datasets.
    \item The performance of the AdaSemSeg evaluated on \emph{unseen} target datasets is comparable to the baselines that are trained only on samples in the target datasets.
    \item The AdaSemSeg comprehensively outperforms the prototype-based FSSS method and the segmentation model trained using transfer learning.
\end{itemize}

\begin{figure*}[hb]
\centering
\includegraphics[width=0.95\textwidth]{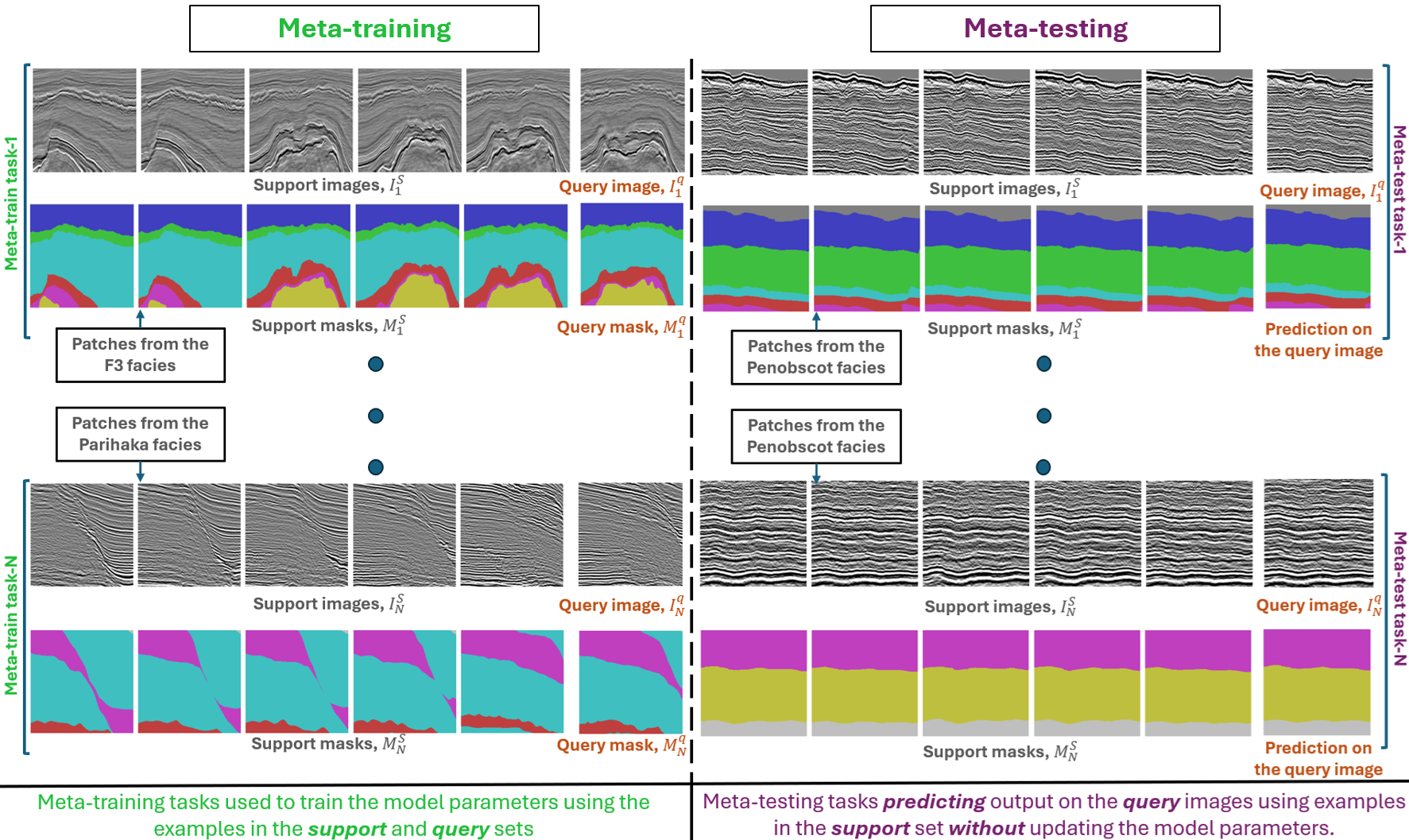}
\caption{Tasks were generated using samples in the seismic datasets for the meta-training and meta-testing stages in the FSSS methods. $I^S$ and $M^S$ represent the support images and corresponding masks used in both the meta-training and meta-testing stages. Similarly, images in the query set ($Q$) are represented as $I^q$ and the associated masks as $M^q$. In this illustration, we use $K=5$ support images and $1$ query images from the F3 \cite{f3_facies_netherlands_2019} and Parihaka \cite{Parihaka_facies_new_zealand_2020} facies datasets in the meta-training stage. A similar setting is used for the Penobscot \cite{Penobscot_facies_canada_2021} dataset in the meta-testing stage. Tasks in the meta-training stage are used to train the parameters of the FSSS model, and the trained model is evaluated on tasks in the meta-testing stage. 
} \label{fig:Meta-train-test}
\end{figure*}

\section{Related Work}
Deep learning methods have been used to recognize geological features, such as faults and channels, in seismic images \cite{f3_faultlines_transferlearn_2019,multitask_2023}. With the release of multiple seismic facies datasets in the recent past \cite{f3_facies_netherlands_2019,Penobscot_facies_canada_2021,Parihaka_facies_new_zealand_2020}, deep learning algorithms have been proposed for the identification of seismic facies \cite{seismic_facies_2022,parihaka_facies_2022,Deeplab_2023}. Among the public datasets, the data released as a part of a SEG challenge on the Parihaka 3D volume \cite{parihaka_facies_2022} has gained much attention due to its complexity. Different segmentation networks, such as the DeconvNet \cite{f3_facies_netherlands_2019}, U-Net \cite{seismic_facies_2022,parihaka_facies_2022}, and DeepLabv3 \cite{Deeplab_2023} have been used for the interpretation of facies.

All the segmentation methods developed for the identification of seismic facies \cite{f3_facies_netherlands_2019,penobscot_danet_2018,seismic_facies_2022,parihaka_facies_2022,Deeplab_2023} rely on a large set of annotated data, which does not present a realistic scenario as the production and annotation of seismic images are costly enterprise. In reality, we would use a pretrained model (trained on benchmark datasets) to make predictions on a new dataset with a handful of annotations. Few-shot segmentation (FSS) is an effective technique used to address these scenario \cite{shaban2017,dgpnet_2022}, which has been extended to multi-class classification \cite{FSSS_ICIP_2022,FSSS_CVPRW_2023}. Among the different FSS methods, we find the DGPNet \cite{dgpnet_2022} using Gaussian processes (GP) in the latent space to be an effective method for segmenting seismic facies. The GP is a probabilistic regression technique that can quickly adapt to the observed data.

FSSS methods have been developed for the segmentation of salt bodies \cite{SAM_Salt_Bodies_1,SAM_Salt_Bodies_2} and seismic facies \cite{FSSS_Facies,facies_selfsupervised_2023,facies_protoseg_2023}. The methods for the segmentation of salt bodies in \cite{SAM_Salt_Bodies_1,SAM_Salt_Bodies_2} use Segment Anything model \cite{SAM_ICCV_2023} trained with over $1$ billion segmentation masks. The method in \cite{FSSS_Facies} adapts the few-shot segmentation method proposed in \cite{Texture_2021} to segment facies in the F3 \cite{f3_facies_netherlands_2019} and Parihaka \cite{parihaka_facies_2022} datasets. Prototype-based few-shot segmentation method is proposed in \cite{facies_protoseg_2023} to identify seismic facies in the F3 \cite{f3_facies_netherlands_2019} and Penobscot \cite{Penobscot_facies_canada_2021} datasets. The FSSS method proposed in \cite{facies_protoseg_2023} for segmenting facies is the closest to our approach as it has the flexibility to deal with multiple datasets varying in the number of classes. Therefore, we compare the proposed AdaSemSeg with the method in \cite{facies_protoseg_2023}.

\begin{figure*}[hb]
\centering
\includegraphics[width=0.90\textwidth]{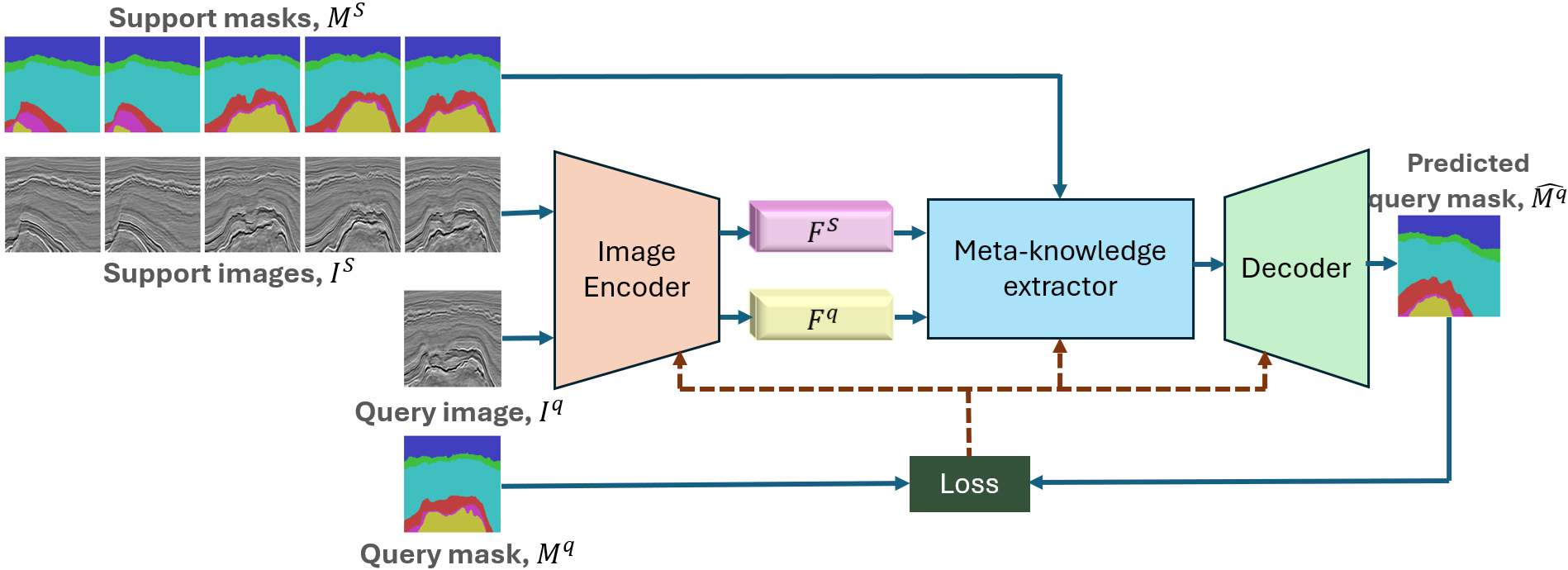}
\caption{A schematic of the few-shot semantic segmentation (FSSS) algorithm. We show here how a task in the meta-training stage is used to train the model parameters. The FSSS method uses $5$ support examples (i.e., $K=5$ shots) from the F3 facies dataset \cite{f3_facies_netherlands_2019} to make \emph{multi-class} prediction (six classes) on the query image. The \textcolor{blue}{forward passes} in the model are represented using \textcolor{blue}{blue arrows}. The predicted mask is compared with the ground truth to compute the loss that is used to update the model parameters. The \textcolor{red}{backward passes} allowing the flow of gradients are represented with \textcolor{red}{red dashed arrows}.} \label{fig:Meta-Learning}
\end{figure*}

\section{Overview of Few-Shot Semantic Segmentation}\label{sec:FSSS}
Few-shot learning is a meta-learning algorithm \cite{MAML_2017,Proto_2017} whose objective is to learn a shared representation from several related tasks, such as the segmentation of images. Few-shot learning used for the segmentation of images is known as the Few-Shot Semantic Segmentation. There are two stages in few-shot learning: the \emph{meta-training} and \emph{meta-testing}. In the meta-training stage, large annotated data, known as the \emph{source} data, are used to train the parameters of a deep neural network to learn generalizable features from different segmentation tasks. The meta-testing stage involves adapting the meta-trained model parameters to new tasks on the \emph{target} data using a handful of annotated samples. However, the trainable model parameters are not fine-tuned with a few annotated examples from the target data. In the context of semantic segmentation of seismic facies studied in this work, we consider one class as the target data and the remaining classes as the source data. For example, the F3 \cite{f3_facies_netherlands_2019} and Parihaka \cite{Parihaka_facies_new_zealand_2020} facies datasets are considered as the source data, where the Penobscot \cite{Penobscot_facies_canada_2021} facies dataset is the chosen as the target data. Therefore, the source and target data contain examples from completely different classes without overlap.

Each task in the meta-training and meta-testing stages is defined using \textit{support} and \textit{query} examples. The support and query examples are selected following the \textit{N}-way \textit{K}-shot structure, where \textit{N} and \textit{K} represent the number of classes and examples (or shots) from each class, respectively. For the segmentation tasks studied in this chapter, we use $N=1$ as in \cite{shaban2017}, i.e., for a task, we select a class at random and sample \textit{K} unique examples that together with the corresponding segmentation masks form the {support} set, represented as $S$. Typically, a single sample from the same class and its segmentation mask is used to construct the {query} set, represented as $Q$. In this work, we denote the \textit{support} set as $S=\{(I^s,M^s)^{i}\}_{i=1}^{K}$ and the query set as $Q=\{(I^q, M^q)\}$, where $I \in R^{H \times W \times 3}$ is the \textit{support/query} image and $M \in \{1, \ldots, C\}^{H \times W}$ is the corresponding segmentation mask with $C$ classes. The value of $C$ is six, seven, and six for the F3 \cite{f3_facies_netherlands_2019}, Penobscot \cite{Penobscot_facies_canada_2021} and Parihaka \cite{Parihaka_facies_new_zealand_2020} facies datasets. Unlike the regular setting of the FSSS methods, the number of classes ($C$) varies across datasets in our application.

Different tasks produced following the \textit{1}-way \textit{5}-shot structure in the meta-training and meta-testing stages using samples extracted from the datasets studied in this work are shown in Figure \ref{fig:Meta-train-test}. We denote the images in the support set ($S$) as $I^S$ and the corresponding masks as $M^S$. Similarly, the images in the query set ($Q$) are represented as $I^q$ and the associated masks as $M^q$. Meta-training tasks are used to train the FSSS model in the meta-training stage. We evaluate the trained model on \emph{unseen} classes in the target data in the meta-testing stage. The meta-testing tasks are also generated using the \textit{1}-way \textit{5}-shot setting used in the meta-training stage for ensuring consistency between the training and evaluation scenarios.

A general layout of the meta-learning algorithm for the FSSS is presented in Figure \ref{fig:Meta-Learning}. In addition to an image encoder and a decoder used in a standard segmentation network, the meta-learning algorithm uses a \emph{meta-knowledge extractor} that learns general visual features from different meta-training tasks (refer to Figure \ref{fig:Meta-train-test}) to solve unseen meta-testing tasks with only \textit{K} annotated examples. In Figure \ref{fig:Meta-Learning}, we illustrate the training of a meta-learning algorithm using a $5$-shot setup, which is the meta-training stage. The predicted segmentation mask is compared with the ground truth to compute the loss that is used to update the model parameters. Under this setting, we train the segmentation model end-to-end using stochastic gradient descent. The flow of gradients is highlighted with red dashed lines. The model parameters are trained on several meta-training tasks as shown in Figure \ref{fig:Meta-train-test} that helps in learning a generalized feature representation.

In Figure \ref{fig:Meta-Learning}, a shared image encoder is used to extract features from the support and query examples for further processing. The initialization of the image encoder using the statistics of a large image dataset, such as the ImageNet \cite{ImageNet_2009}, plays an important role in the performance of the FSSS. This is conceivably due to the learning of rich visual features from a large dataset with a lot of variability that assists in solving other recognition tasks \cite{Imagenetgoodtransfer}. In general, the FSSS algorithm freezes the update of the model parameters initialized with the statistics of a large dataset, and this strategy has been found to be effective for making predictions on unseen classes in the meta-testing stage. However, the availability of a vast annotated dataset is challenging in many applications, such as the interpretation of seismic images. Under such circumstances, contrastive self-supervised algorithms, such as the SimCLR \cite{SimCLR_2020}, offer an alternative approach to initialize the backbone image encoder. The advantage of self-supervised algorithms is that we \emph{do not need labeling of the data}, unlike the ImageNet \cite{ImageNet_2009}. Contrastive self-supervised algorithms use positive and negative examples produced from \emph{unlabeled samples} in a dataset to train the network parameters.

\begin{figure*}[ht]
\centering
\includegraphics[width=0.90\textwidth]{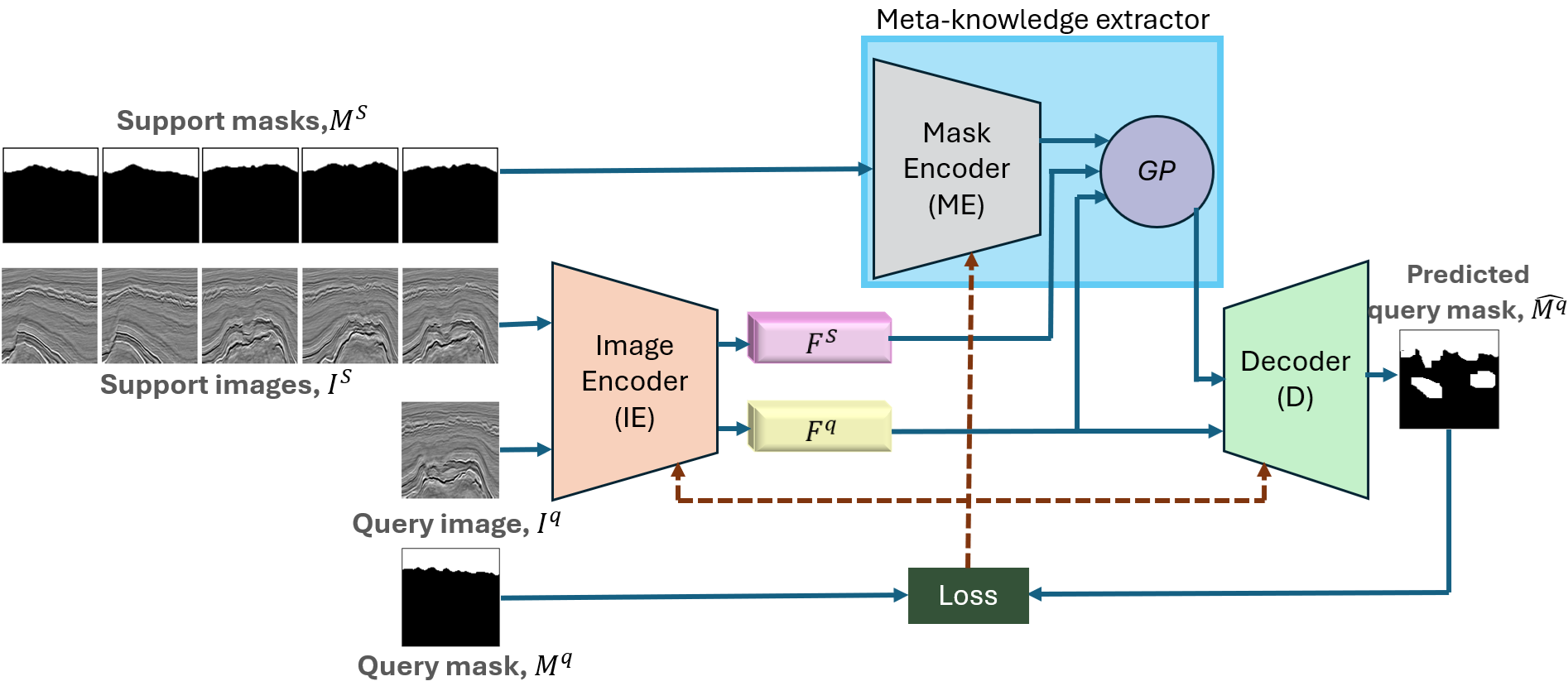}
\caption{The architecture of the DGPNet \cite{dgpnet_2022} used for the binary segmentation task. The DGPNet uses $K=5$ support examples to predict the binary segmentation mask for a single facies type in the F3 facies dataset \cite{f3_facies_netherlands_2019} on the query image. The \textcolor{blue}{forward passes} in the model are represented using \textcolor{blue}{blue arrows}. The predicted mask is compared with the ground truth to compute the loss that is used to update the model parameters. The \textcolor{red}{backward passes} allowing the flow of gradients are represented with \textcolor{red}{red dashed arrows}.} \label{fig:DGPNet}
\end{figure*}

\begin{figure*}[bh]
\centering
\includegraphics[width=1.0\textwidth]{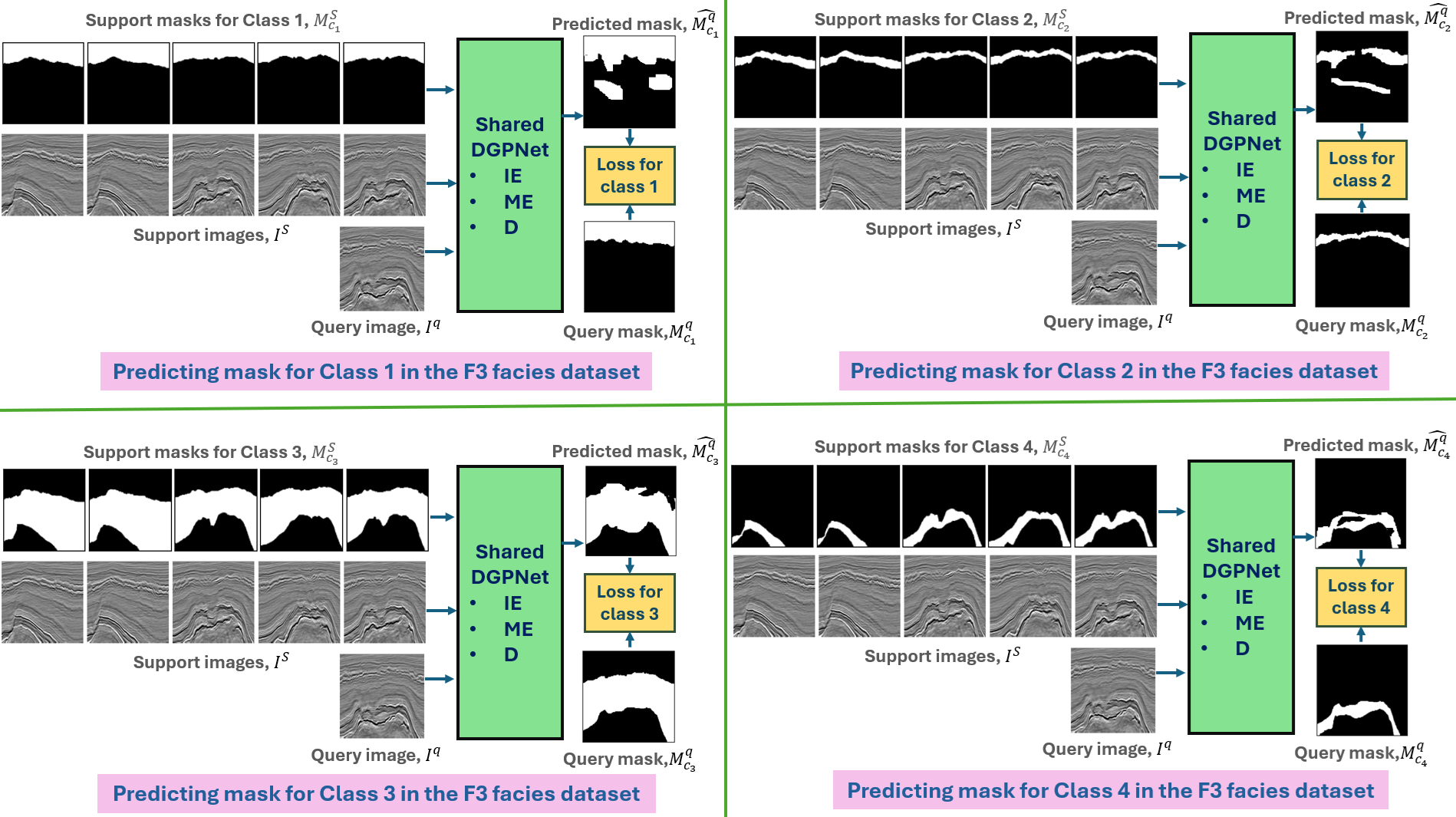}
\caption{Here we show the use of a \emph{shared} DGPNet \cite{dgpnet_2022} for the binary segmentation task of \emph{four} different types of facies in the F3 facies dataset \cite{f3_facies_netherlands_2019}. In each quadrant, the \emph{same} DGPNet uses $K=5$ support examples to predict the binary segmentation mask for a specific facies type on the query image. The forward passes in the model are represented using \textcolor{blue}{blue arrows}. The class-specific predicted masks are compared with the corresponding ground truths to compute class-wise losses.} \label{fig:DGPNet_Multiclass}
\end{figure*}

\begin{figure*}[th]
\centering
\includegraphics[width=0.8\textwidth]{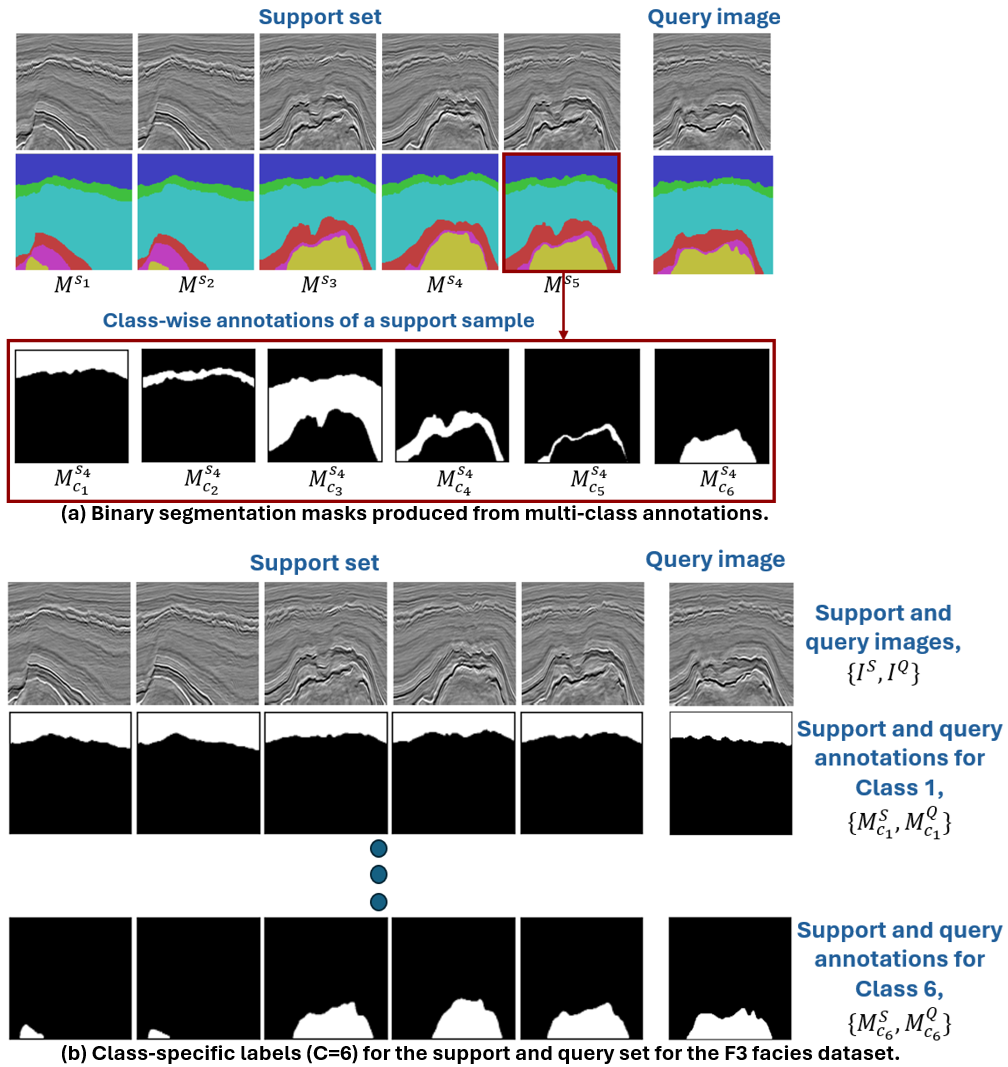}
\caption{The processing of the input to produce class-specific annotations on the support and query images. (a) Processing of the multi-class labeled example, $M^{s_5}$, in the support set to produce binary segmentation masks for $6$ classes in the F3 dataset \cite{f3_facies_netherlands_2019}. (b) Class-specific labels for the support ($S$) and query set ($Q$) for class $c_1$ and class $c_6$ of the F3 facies dataset \cite{f3_facies_netherlands_2019}.} \label{fig1:input_processing}
\end{figure*}

\subsection{Segmentation using the DGPNet \cite{dgpnet_2022}}\label{subsec:DGPNet}
In this section, we discuss the formulation of the DGPNet \cite{dgpnet_2022}
developed for \emph{binary segmentation} (a two-class segmentation task) of natural image datasets. 
We adapt the DGPNet in the proposed AdaSemSeg to segment seismic facies. The DGPNet's novelty is in using GP regressions in deep latent spaces. We present a schematic of the DGPNet \cite{dgpnet_2022} in Figure \ref{fig:DGPNet} demonstrating the binary segmentation of a single facies type in the F3 facies dataset. The DGPNet \emph{introduces a mask encoder for GP regressions in the latent layers}, and we leverage this design choice in our proposed method, AdaSemSeg, to address the variability in the number of classes across datasets. The GP regressions in the latent layers of the DGPNet predict the segmentation mask in deeper layers, which, combined with the encoded image features in shallow layers, predicts the final segmentation output. We discuss details in subsequent paragraphs.

The DGPNet, as shown in Figure \ref{fig:DGPNet}, has three trainable modules, namely, the image encoder (IE), mask encoder (ME), and decoder (D). The image encoder encodes images in the support and query set, whereas the mask encoder encodes the binary masks in the support set. The decoder processes the output of GP regressions in the latent space and the shallow encoded image features to predict binary masks. We will use the terminologies introduced in Section \ref{sec:FSSS} for explaining binary few-shot segmentation using the DGPNet. Using the support images, $I^S$, and the corresponding binary masks, $M^S$, in the support set, the segmentation network predicts a binary mask, $\hat{M^q}$, on the query image, using the prediction of GP regression (in deep layers) and the encoded query image features (in the shallow layers). The predicted binary mask, $\hat{M^q}$, is compared with the ground truth $M^q$ to minimize the mismatch. Under this setting, we train the segmentation model end-to-end using stochastic gradient descent. We train the DGPNet on several meta-training tasks from the source data that help learn robust and generalizable representations. The learned representations are used in the meta-testing stage to predict the masks on samples in the target data.

In this paragraph, we explain the details of GP regressions used in the latent layers of the DGPNet. GP regression in the latent layer learns the mapping from the encoded image space to the encoded mask space. The encoded representation of a support image, $I^s$, and its corresponding mask, $M^s$, are denoted as $e^s = {\rm IE}(I^s) \in R^{H^{'} \times W^{'} \times F}$ and $e^{s^{'}} = {\rm ME}(M^s) \in R^{H^{'} \times W^{'} \times F^{'}}$, respectively. $KH^{'}W^{'}$ image and mask encodings are produced from $K$ samples in the support set, $S=\{(I^s,M^s)^{i}\}_{i=1}^{K}$, in the $F$ and $F^{'}$ dimensional feature space, respectively. Similarly, the encoded representation of a {query} image, $I^q$, defined as $e^q = {\rm IE}(I^q) \in R^{H^{'} \times W^{'} \times F}$ results in $H^{'}W^{'}$ examples in the encoded image space. GP regression in the latent layers of the DGPNet learns the mapping from the encoded image space, $R^{F}$, to the encoded mask space, $R^{F^{'}}$, using $KH^{'}W^{'}$ image and mask encodings derived from the support set, $S$. GP regression is a probabilistic regression technique, and the statistics of the posterior distribution, $P=\{\mu_{q|S}, \Sigma_{q|S}\}$, estimated by the regression model \cite{GP_book} are defined as,
\begin{equation}\label{eq:mean_GP}
\mu_{q|S} = K_{Sq}^{T}(K_{SS} + \sigma_{z}^{2}{\mathbf{I}})^{-1}e^{S^{'}} \in R^{H^{'} \times W^{'} \times F^{'}}
\end{equation}
\begin{equation}\label{eq:cov_GP}
\Sigma_{q|S} = K_{qq} - K_{Sq}^{T}(K_{SS} + \sigma_{z}^{2}{\mathbf{I}})^{-1}K_{Sq} \in R^{H^{'}W^{'} \times H^{'}W^{'}},
\end{equation}
where $K_{SS}$, $K_{qq}$, and $K_{Sq}$ are the co-variance matrices computed using the encoded support images, $e^S=\{e^{s_i}\}_{i=1}^{K}$, and encoded query image, $e^q$. The noise in the labeled data is represented by $\sigma_{z}$. A squared exponential kernel (${k}_{\rm SE}$) used for computing the co-variance matrices, $K_{SS}$, $K_{qq}$, and $K_{Sq}$, is defined as follows:
\begin{equation}\label{eq:squared_exp_kernel}
{k}_{\rm SE}(z_1, z_2) = \sigma^{2}\exp{\frac{-\|z_2-z_1\|_{2}^{2}}{2l^{2}}},
\end{equation}
where $l$ and $\sigma$ represent the kernel bandwidth and scaling factor, respectively.

The DGPNet effectively segments natural images using off-the-shelf segmentation models, such as the ResNet \cite{ResNet_2016} as an image encoder and the DFN \cite{DFN_DEcoder_DGPNet_2018} as the decoder. We can change the backbone networks of the DGPNet depending on the application. For example, our work does not use the DFN as the decoder for segmenting seismic facies. We observed that the performance of the DGPNet is strongly impacted by the initialization of the image encoder with the pretrained statistics of the large and complex ImageNet dataset \cite{ImageNet_2009}. This is because the ImageNet statistics help with the prior knowledge of almost all natural objects likely to be encountered as unseen classes in the target data. This demonstrates the importance of the initialization of the image encoder on the performance of the FSSS methods. We address this issue in our application using a different initialization technique.

The existing formulation of the DGPNet can be extended to solve a multi-class segmentation problem, such as the segmentation of seismic facies. However, similar to other FSSS methods, we need to fix the number of output classes in the decoder network. Therefore, we cannot use a DGPNet trained on the F3 or Parihaka facies dataset, both having six facies types, to make inferences on samples in the Penobscot dataset having seven different types of facies. We solve this problem using the proposed AdaSemSeg, which can handle the variability in the number of classes across datasets.

\section{The Proposed Method: AdaSemSeg} \label{sec:method}
In this section, we present the proposed AdaSemSeg, a few-shot semantic segmentation method developed for the segmentation of seismic facies (a multi-class segmentation problem) that can handle the variability in the number of classes across datasets. To deal with the variability in the number of classes across datasets, the AdaSemSeg splits the original multi-class segmentation problem into several binary segmentation problems. The AdaSemSeg uses the DGPNet \cite{dgpnet_2022} as the backbone to solve the binary segmentation tasks. The parameters of the DGPNet are \emph{shared} across all the binary segmentation tasks in the AdaSemSeg. Therefore, the number of \emph{trainable parameters in the proposed method do not grow with the number of classes in a dataset}. Similar to other FSSS methods, the model parameters are trained on the source data in the meta-training stage. The trained parameters are used in the meta-testing stage to predict segmentation masks of samples in the target data using a few annotated examples from the same dataset. We will use the terminologies introduced in Section \ref{sec:FSSS} for explaining the AdaSemSeg.

We need to fix the number of output classes in the decoder of the existing few-shot segmentation methods as deep neural networks learn a deterministic mapping from the input image to output segmentation masks. The use of GP regression (a probabilistic regression method) in the DGPNet offers the flexibility to solve different segmentation tasks using the same support images ($I^S$) based on the segmentation masks ($M^S$) fed as input to the mask encoder (ME). Therefore, the DGPNet produces different predictions on a query image based on the input to the ME. The mean $\mu_{q|S}$ of GP regression defined in equation \ref{eq:mean_GP} is a function of the encoded support mask ($e^{S^{'}} = {\rm ME}(M^S)$), which explains the prediction of different binary masks on the same query image based on the support masks, $M^S$. In Figure \ref{fig:DGPNet_Multiclass}, we demonstrate the prediction of the \emph{different segmentation masks} on the \emph{same query image} by the DGPNet depending on the support masks provided as input to the ME. In this example, we predict four different types of facies (out of six) in the F3 facies dataset using the same parameter settings of the DGPNet, i.e., the DGPNet is shared across all the binary segmentation tasks. The use of different segmentation masks, $\{M^S_{c_1}$, $M^S_{c_2}$, $M^S_{c_3}$, $M^S_{c_4}\}$, for the same support images, $I^S$, leads to the prediction of different segmentation masks, $\{\hat{M^q_{c_1}}, \hat{M^q_{c_2}}, \hat{M^q_{c_3}}, \hat{M^q_{c_4}}\}$ on the query image, $I^q$. We identified this property in the DGPNet and used it in the AdaSemSeg to deal with the variability in the number of facies across seismic datasets.

\subsection{Processing of data for binary segmentation tasks}\label{subsec:data_processing}
As discussed in the previous paragraph, the AdaSemSeg segments seismic facies, a multi-class segmentation problem, by solving several binary segmentation tasks using a shared DGPNet \cite{dgpnet_2022}. To this end, the original labeled data, $M^s$, having multi-class annotations, is processed to produce binary segmentation masks, $M_c^s$, representing the mask for class $c$, as shown in Fig. \ref{fig1:input_processing}(a) for an example in the support set. The number of binary masks produced for a labeled input depends on the number of facies, $C$, in the dataset, resulting in $M^{s_i}=\{M_{c_j}^{s_i}\}_{j=1}^{C}$ for a labeled mask $M^{s_i}$ in the support set. We produce \emph{class-specific annotations} support and query set using this processing step, defined as $M^S_{c_{j}}=\{M_{c_{j}}^{s_{i}}\}_{i=1}^{K}$ and $M^Q_{c_{j}}=\{M_{c_{j}}^q\}$, respectively, for class ${c_j}$. Fig. \ref{fig1:input_processing}(b) shows the class-specific labels, $\{M^S_{c_1}, M^Q_{c_1}\}$ and $\{M^S_{c_6}, M^Q_{c_6}\}$ for the support and query set, representing class $c_1$ and $c_6$ of the F3 facies dataset, respectively. Processing of the input using this technique results in multiple class-specific annotations, depending on the number of facies ($C$), for the same images ($I$) in the support and query set. The processing of the annotations in the support and query set to produce binary masks is performed \emph{on the fly} during the training/inference of the AdaSemSeg.

\begin{figure*}[bh]
\includegraphics[width=0.97\textwidth]{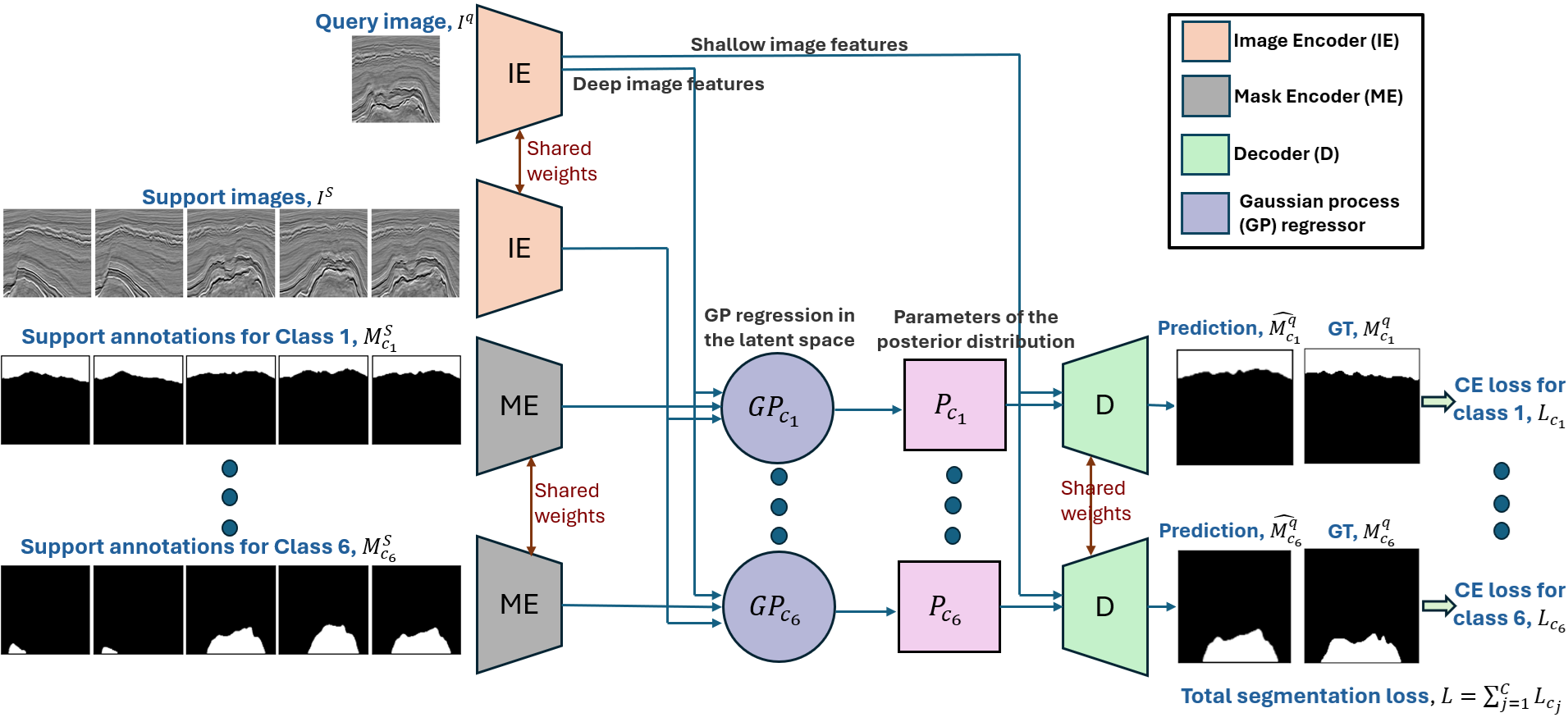}
\caption{Training of the AdaSemSeg using Algorithm \ref{alg:Meta-train-AdaSemSeg} on the F3 facies dataset \cite{f3_facies_netherlands_2019} in the $5$-shot setup. The AdaSemSeg predicts class-wise binary segmentation masks using a shared DGPNet\cite{dgpnet_2022}. \textcolor{red}{\emph{Shared weights}} indicates the \textcolor{red}{\emph{same}} neural network parameters used to process different inputs, such as the same mask encoder (ME) is used to encode class-specific binary masks, $M^S_{c_{j}}$, in the support set. The forward passes are represented using \textcolor{blue}{blue arrows}.} \label{fig:AdaSemSeg}
\end{figure*}

\subsection{Network architecture}\label{subsec:network_arch_AdaSemSeg}
The AdaSemSeg segments seismic facies by solving multiple class-wise binary segmentation tasks using \emph{a single shared} DGPNet \cite{dgpnet_2022}. Therefore, the number of trainable parameters in the AdaSemSeg is the same as the DGPNet. Similar to the DGPNet, the AdaSemSeg has three trainable modules, namely, the image encoder (IE) that encodes images in the support and query set, the mask encoder (ME) that encodes the class-specific binary masks, $M^S_{c_{j}}$, in the support set and, the decoder (D) that processes the output of the GP regressions in the latent space and the shallow encoded image features to predict binary masks, as shown in Fig. \ref{fig:AdaSemSeg}. The IE, ME, and D are deep neural networks parameterized by $\phi$, $\psi$, and $\theta$, respectively. In Fig. \ref{fig:AdaSemSeg}, \emph{shared weights} indicates the same neural network parameters used to process different inputs. For example, the same mask encoder is used to encode $K$ class-specific binary masks in the support set $M^S_{c_{j}}=\{M_{c_{j}}^{s_{i}}\}_{i=1}^{K}$. Therefore, the number of trainable parameters in the AdaSemSeg is \emph{fixed} and \emph{does not grow} with the number of facies in a dataset. This example demonstrates the segmentation of facies in the F3 \cite{f3_facies_netherlands_2019} dataset.

The AdaSemSeg uses the support images, $I^S$, and class-specific binary masks in the support set, $M^S_{c_j}$ for the class $c_j$, to predict a binary mask, $\hat{M^q_{c_j}}$, on the query image. The predicted binary mask, $\hat{M^q_{c_j}}$, is produced using the output of GP regression (in deep layers) and the encoded query image features (in the shallow layers). The GP in a latent layer in the AdaSemSeg learns the mapping from the encoded image space to the encoded mask space, similar to the DGPNet. The statistics of the class-wise posterior distribution, $P_{c_j}=\{\mu_{q_{c_j}|S_{c_j}}, \Sigma_{q_{c_j}|S_{c_j}}\}$, of GP regression \cite{GP_book} are defined as,
\begin{align}
\mu_{q_{c_j}|S_{c_j}} &= K_{Sq}^{T}(K_{SS} + \sigma_{z}^{2}{\mathbf{I}})^{-1}e_{c_j}^{S^{'}} \in R^{H^{'} \times W^{'} \times F^{'}} \label{eq:AdaSemSegGP_mean} \\
\Sigma_{q_{c_j}|S_{c_j}} &= K_{qq} - K_{Sq}^{T}(K_{SS} + \sigma_{z}^{2}{\mathbf{I}})^{-1}K_{Sq} \in R^{H^{'}W^{'} \times H^{'}W^{'}},\label{eq:AdaSemSegGP_cov}
\end{align}
where $K_{SS}$, $K_{qq}$, and $K_{Sq}$ are the co-variance matrices computed using the encoded support images, $e^S=\{e^{s_i}\}_{i=1}^{K}$, and encoded query image, $e^q$. The details of GP regression are reported in Section \ref{subsec:DGPNet}. The expressions in equation \ref{eq:AdaSemSegGP_mean} and \ref{eq:AdaSemSegGP_cov} hold the same meaning as defined in Section \ref{subsec:DGPNet}. It should be noted that the $\mu_{q_{c_j}|S_{c_j}}$ in equation \ref{eq:AdaSemSegGP_mean} is a function of the class-specific mask encoding, $e_{c_j}^{S^{'}}$, and the $\Sigma_{q_{c_j}|S_{c_j}}$ in equation \ref{eq:AdaSemSegGP_cov} is independent of a class.

\begin{algorithm}[t]
\caption{: \textbf{Meta-training of the AdaSemSeg}}
\begin{algorithmic}
    \STATE {\bfseries Input:} $\mathcal{X}_{\rm src}$, $batchSize$, $K$
    \STATE {\bfseries Output:} Trained image encoder, mask encoder and decoder parameters represented by $\phi$, $\psi$, and $\theta$, respectively.
    \STATE \medskip \COMMENT{\textbf{Split data and initialize trainable parameters}}
    \STATE Split $\mathcal{X}_{\rm src}$ into training, $\mathcal{X}_{\rm src}^{\rm train}$, and validation data, $\mathcal{X}_{\rm src}^{\rm val}$
    \STATE $\mathcal{X}_{\rm src}^{\rm val}$ is used for adjusting the learning rate
    \STATE Initialize $\phi$, $\psi$ and $\theta$
    \STATE \medskip \COMMENT{\textbf{Meta-training over an epoch}}
    \FOR {number of minibatch updates}
    \STATE \medskip \COMMENT{\textbf{Minibatch from the training set with $K-$shots}}
    \STATE miniBatch = getMiniBatch($\mathcal{X}_{\rm src}^{\rm train}, batchSize, K)$
    \STATE \medskip \COMMENT{\textbf{Loss over a minibatch}}
    \STATE totalLoss = 0
    \FOR {b in $[0, 1, 2, \ldots, \textit{batchSize}-1]$}
    \STATE $C$, $I^S$, $M^S$, $I^q$, $M^q$ = miniBatch[b]
    \STATE \medskip \COMMENT{\textbf{Get the class-specific binary masks}}
    \STATE $\{M_{c_j}^{S}\}_{j=1}^{C}$, $\{M_{c_j}^{q}\}_{j=1}^{C}$ = getBinaryMask($M^S$, $M^q$, $C$)
    \STATE \medskip \COMMENT{\textbf{Accumulates the loss over all the classes}}
    \STATE classwiseLoss = 0
    \FOR {j in $[1, 2, \ldots, C]$}
    \STATE \medskip \COMMENT{\textbf{Shared parameters of the DGPNet}}
    \STATE $\hat{M^q_{c_j}}$ = DGPNet($I^S$, $M_{c_j}^{S}$, $I^q$)
    \smallskip
    \STATE classwiseLoss += CrossEntropy($\hat{M^q_{c_j}}$, $M_{c_j}^{q}$)
    \ENDFOR
    \medskip
    \STATE totalLoss += classwiseLoss
    \ENDFOR
    \STATE \medskip \COMMENT{\textbf{Update the parameters of the DGPNet}}
    \STATE Update $\phi$, $\psi$ and $\theta$ by minimizing the total segmentation loss, totalLoss, using Stochastic Gradient Descent
    \smallskip
    \ENDFOR
    \end{algorithmic}
\label{alg:Meta-train-AdaSemSeg}
\end{algorithm}

\begin{algorithm}[t]
\caption{: \textbf{Meta-testing of the AdaSemSeg}}
\begin{algorithmic}
    \STATE {\bfseries Input:} $\mathcal{X}_{\rm target}$, $K$, $C$, Trained AdaSemSeg parameters
    \STATE {\bfseries Output:} Predictions on samples in $\mathcal{X}_{\rm target}$ using only $K$ annotated samples
    \STATE \medskip \COMMENT{\textbf{$K$ support images in the target data with $C$ classes}}
    \STATE $S=\{(I^s,M^s)^{i}\}_{i=1}^{K}$
    \STATE \medskip \COMMENT{\textbf{We treat the remaining images as the query set}}
    \STATE Q = $\mathcal{X}_{\rm target}$ - S
    \STATE \medskip \COMMENT{\textbf{Predictions on the query set}}
    \STATE results = []
    \FOR {$I^q$, $M^q$ in $Q$}
    \STATE \medskip \COMMENT{\textbf{Get the class-specific binary masks}}
    \STATE $\{M_{c_j}^{S}\}_{j=1}^{C}$ = getBinaryMask($M^S$, $C$)
    \STATE \medskip \COMMENT{\textbf{Class-wise predictions on the query image}}
    \STATE classwisePreds = []
    \FOR {j in $[1, 2, \ldots, C]$}
    \STATE \medskip \COMMENT{\textbf{Predictions using the trained AdaSemSeg}}
    \STATE \COMMENT{\textbf{Using the AdaSemSeg in the evaluation mode}}
    \STATE $\hat{M^q_{c_j}}$ = DGPNet($I^S$, $M_{c_j}^{S}$, $I^q$)
    \smallskip
    \STATE classwisePreds[j-1] = $\hat{M^q_{c_j}} \in R^{H \times W}$
    \smallskip
    \ENDFOR
    \STATE \medskip \COMMENT{\textbf{Get the class index with maximum probability}}
    \STATE $\hat{M^q} = \argmax\limits_{C}(\text{classwisePreds})  \in R^{H \times W}$
    \STATE results.append($I^q$, $M^q$, $\hat{M^q}$)
    \smallskip
    \ENDFOR
    \smallskip 
    \STATE return results
    \end{algorithmic}
\label{alg:Meta-test-AdaSemSeg}
\end{algorithm}

\subsection{Meta-training}\label{subsec:train_AdaSemSeg}
In the meta-training stage, we train the parameters of the AdaSemSeg on a large annotated source data, $\mathcal{X}_{\rm src}$, to learn generalizable features. This work uses two benchmark datasets (out of three) as the source data. Independent segmentation tasks are constructed from the source data following the \textit{N}-way \textit{K}-shot structure as done in any few-shot semantic segmentation algorithm (refer to Section \ref{sec:FSSS} for more details). Description of the meta-training stage of the AdaSemSeg is reported in Algorithm \ref{alg:Meta-train-AdaSemSeg}.

In the meta-training stage defined in Algorithm \ref{alg:Meta-train-AdaSemSeg}, we split $\mathcal{X}_{\rm src}$ into training, $\mathcal{X}_{\rm src}^{\rm train}$, and validation data, $\mathcal{X}_{\rm src}^{\rm val}$. In our method, the validation data, $\mathcal{X}_{\rm src}^{\rm val}$, is used for adjusting the learning rate of the optimizer. A sample in a minibatch from the training data, $\mathcal{X}_{\rm src}^{\rm train}$, gives us the support ($I^S$, $M^S$) and query ($I^q$, $M^q$) examples from a dataset in the source data along with the number of classes ($C$) in that dataset. In the AdaSemSeg, we divide the multi-class segmentation problem into several binary segmentation problems. Therefore, we process the multi-class segmentation masks to produce class-specific binary masks, $\{M_{c_j}^{S}\}_{j=1}^{C}$, and $\{M_{c_j}^{q}\}_{j=1}^{C}$ corresponding to the support ($M^S$) and query ($M^q$) set, respectively. We follow the technique discussed in section \ref{subsec:data_processing} to produce the class-specific binary masks. Using the class-specific binary masks $M^S_{c_j}$ in the support set, we predict the binary mask on the query image for class ${c_j}$ using the DGPNet \cite{dgpnet_2022}. The predicted mask on the query image for class ${c_j}$, $\hat{M_{c_j}^q}$, is compared with the ground truth annotation, $M_{c_j}^q$ to compute the loss for the class ${c_j}$, $L_{c_j}$ as shown in Fig. \ref{fig:AdaSemSeg}. However, we do not update the parameters of the DGPNet using the class-specific loss $L_{c_j}$. We repeat the above steps to predict the segmentation masks on the query image for the remaining $C-1$ classes in the facies dataset using the class-specific binary masks in the support set. This results in multiple GP regressions, a key difference of the AdaSemSeg with the DGPNet \cite{dgpnet_2022}. It must be noted that we use the same DGPNet parameters across all facies types in a dataset, i.e., the IE, ME, and D are shared across classes.

\begin{figure*}[hb]
\includegraphics[width=\textwidth]{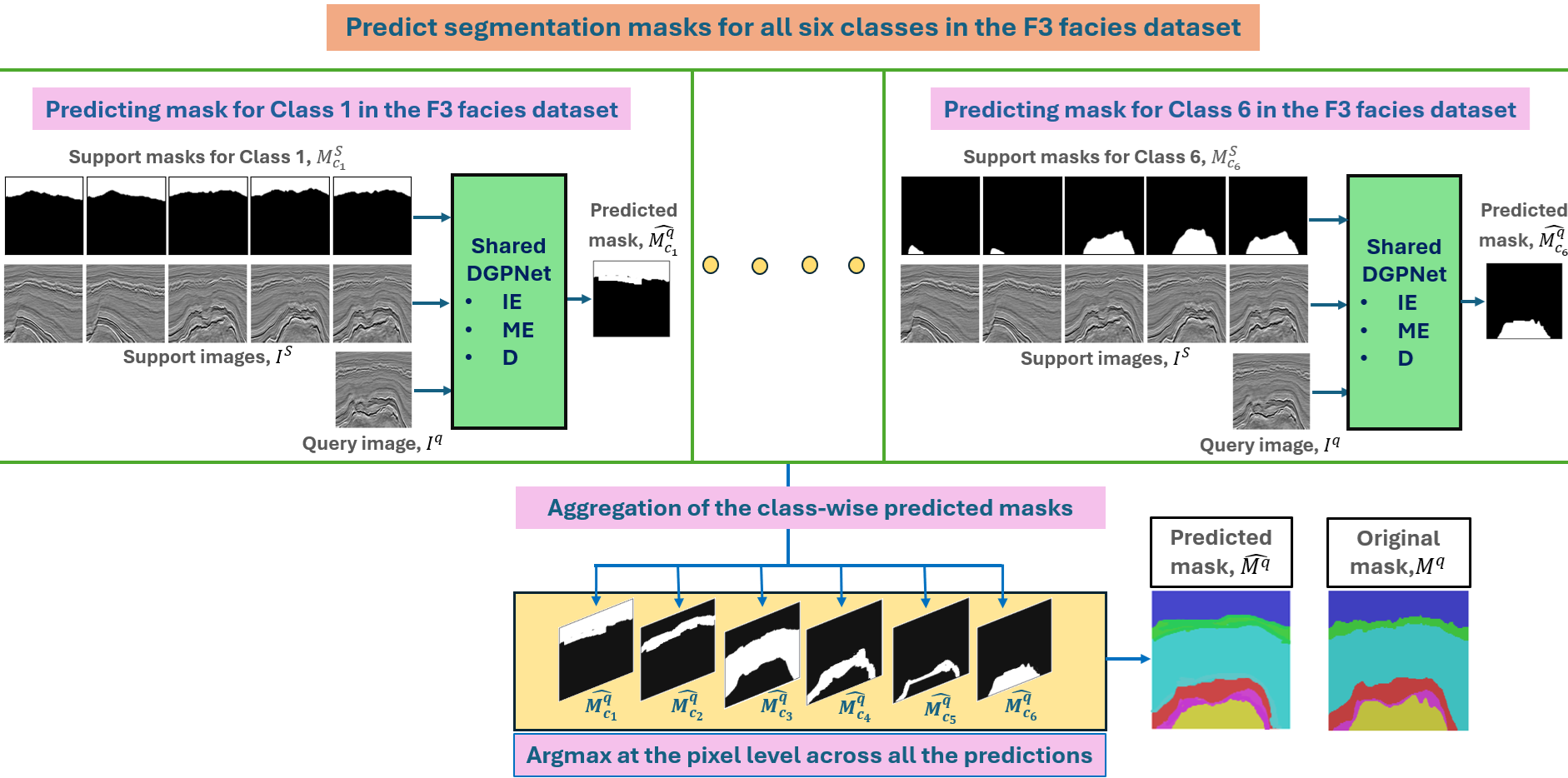}
\caption{Inference using the AdaSemSeg following Algorithm \ref{alg:Meta-test-AdaSemSeg} on the F3 facies dataset \cite{f3_facies_netherlands_2019} in the $5$-shot setup. The AdaSemSeg predicts class-wise binary segmentation masks $\hat{M^q_{c_j}}$ using a \emph{shared} DGPNet\cite{dgpnet_2022}. The forward passes are represented using \textcolor{blue}{blue arrows}. We accumulate the class-wise predicted mask on the query image and get the multi-class prediction $\hat{M^q}$ using the \emph{argmax} at each pixel location.}\label{fig:Inference_AdaSemSeg}
\end{figure*}

The AdaSemSeg accumulates the loss across all the classes, named as the \emph{total segmentation loss}, $ L = \sum_{j=1}^{C}L_{c_j}$. We use the pixel-wise weighted binary cross-entropy loss to compare the predicted masks with the ground truth, and the segmentation loss across multiple classes is defined as,
\begin{equation}\label{eq:CE_AdaSemSeg}
L = -\frac{1}{CHW}\sum_{j=1}^{C}\sum_{h=1}^{H}\sum_{w=1}^{W}w_{c_j}M_{c_j}^{q}(h,w) \log \hat{M_{c_j}^{q}}(h,w).
\end{equation}
The loss function in equation \ref{eq:CE_AdaSemSeg} computes the average pixel-wise weighted binary cross-entropy loss over the number of classes, $C$, in the dataset, where $w_{c_j}$ is the weight of the class $c_j$. The total loss calculated on a minibatch is used to update the shared trainable parameters using stochastic gradient descent. Following the outline in Algorithm \ref{alg:Meta-train-AdaSemSeg}, we train the AdaSemSeg over several epochs until convergence.

\subsection{Meta-testing}\label{subsec:inference_AdaSemSeg}
After the AdaSemSeg is trained on the source data, we evaluate its performance on the target data in the meta-testing stage using a few annotated examples ($K$-shots) from the same dataset. It should be noted that the parameters of the AdaSemSeg are \emph{not fine-tuned} to $K$ annotated examples from the target dataset, $\mathcal{X}_{\rm target}$, and we use the trained AdaSemSeg only for evaluation. Therefore, the meta-testing is equivalent to the inference of a trained deep neural network. To maintain consistency between the training and evaluation of the model, we use the same number of support examples, $K=\{1, 5\}$, in both the meta-training and meta-testing stages. We outline the steps in the meta-testing stage of the AdaSemSeg in Algorithm \ref{alg:Meta-test-AdaSemSeg}. Similar to the meta-training stage, we predict the binary segmentation mask on a query image using $K$ class-specific binary masks in the support set. The prediction of the binary segmentation masks on the query image is repeated for all the $C$ classes in the target dataset. We accumulate the binary predictions on different facies types in a dataset and apply \emph{argmax} at the pixel level to obtain the multi-class prediction on the query image. We repeat this process to evaluate all the test samples in the target dataset. 

In Fig. \ref{fig:Inference_AdaSemSeg}, we illustrate the prediction of the multi-class segmentation mask by the AdaSemSeg on the F3 facies dataset (i.e., the target dataset in this context) using $K=5$ annotated support examples. For each class in the F3 facies dataset, we predict the binary segmentation mask of the query image using the class-specific binary mask of the support examples and a trained DGPNet shared across all classes. We stack the class-wise predicted mask on the query image and get the multi-class prediction using the \emph{argmax} at each pixel location.

\begin{figure*}[th]
\includegraphics[width=\textwidth]{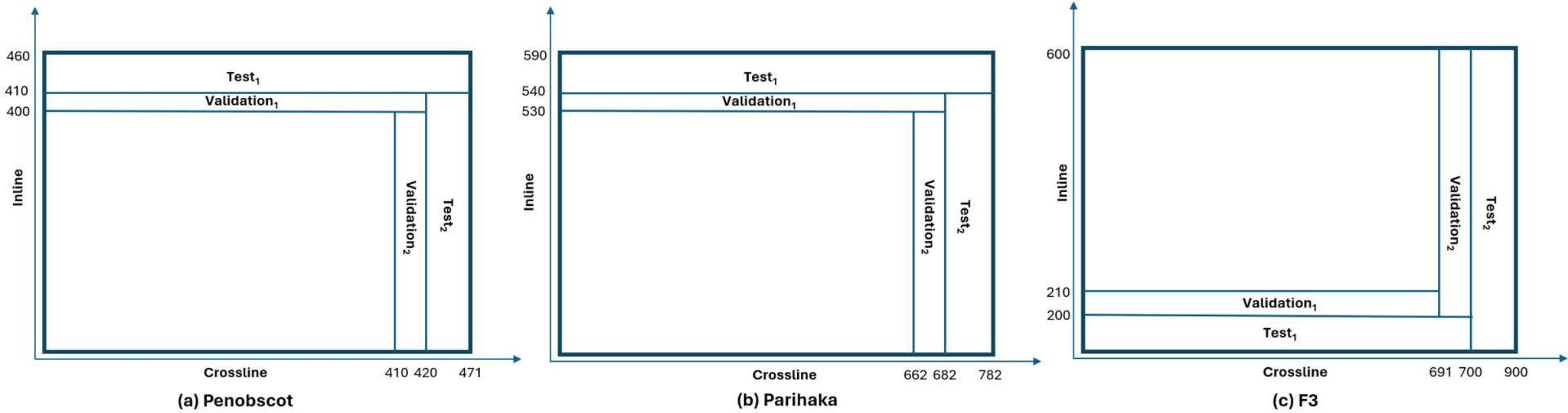}
\caption{The train, validation, and test distribution of the (a) Penobscot \cite{Penobscot_facies_canada_2021}, (b) Parihaka \cite{Parihaka_facies_new_zealand_2020} and (c) F3 \cite{f3_facies_netherlands_2019} datasets.} \label{fig3:Train_val_test_split}
\end{figure*}

\section{Experiments}
\subsection{Experimental Setup}\label{subsec:exp_setup}
\subsubsection{Datasets} We use the Penobscot \cite{Penobscot_facies_canada_2021}, Parihaka \cite{Parihaka_facies_new_zealand_2020}, and F3 \cite{f3_facies_netherlands_2019} facies datasets for evaluation of the AdaSemSeg under different scenarios. The train, validation, and test set distribution for all the datasets are shown in Fig. \ref{fig3:Train_val_test_split}. For the F3 dataset, we use the train-val-test split proposed in \cite{f3_facies_netherlands_2019}. The Penobscot dataset is processed to remove the corrupted images that resulted in 460 inline and 471 crossline images from the original volume containing 601 inline and 481 crossline slices \cite{penobscot_danet_2018}. With the expert's suggestion, we reduced the depth of the Penobscot dataset to $900$ (originally $1501$). We do not merge underrepresented horizons, such as layers 2 and 3, as done in \cite{penobscot_danet_2018}. The Parihaka facies dataset was released as a part of the challenge, where the organizers withheld the original test annotations. Thus, we select a small part of the training data as the test data for evaluating different methods, such that the selected test slices are close to the real test data along the inline and crossline directions \cite{parihaka_facies_2022}. 

We process the Penobscot and Parihaka datasets using percentile-based filtering, such as the $1-99$ percentile, to remove the extreme intensity values (aka \emph{outliers}). We estimate an acceptable range of values $[\textit{lower-threshold}, \textit{upper-threshold}]$ using the percentile-based filtering, such that intensity values less than the $\textit{lower-threshold}$ are clipped to $\textit{lower-threshold}$, and values higher than the $\textit{upper-threshold}$ are mapped to $\textit{upper-threshold}$. This work uses the $5-95$ percentile range to remove the outliers. We re-scale the values of all the datasets to the range of $0-255$. 

We extract 2D patches of size $256 \times 256$ along the inline and crossline directions to train the AdaSemSeg and other methods studied in this work for all the data volumes \cite{f3_facies_netherlands_2019,seismic_facies_2022,facies_protoseg_2023}. Using whole slices instead of patches gives better results for a dataset as demonstrated in \cite{f3_facies_netherlands_2019,parihaka_facies_2022}. However, we hypothesize that patches offer more variability in the training data, resulting in better generalization to other unseen datasets. In addition, whole slices require a lot of GPU memory to train the deep neural networks, which can be impractical for large data volumes. Therefore, we use 2D \emph{patches} of size $256 \times 256$ for all seismic datasets to train the AdaSemSeg and other competing methods studied in this work. However, we evaluate a trained model using \emph{whole slices} from a target dataset.

We use the \emph{leave-one-out} policy to create the data for the meta-training and meta-testing. For example, to evaluate the AdaSemSeg on the Parihaka dataset (target data used in the meta-testing stage), we train the AdaSemSeg on the Penobscot and F3 dataset (source data used in the meta-training stage). Similarly, the AdaSemSeg is evaluated on the Penobscot and F3 datasets when the model is trained on the remaining two datasets. Under this experimental setting, we assess the generalization of the AdaSemSeg to unseen target datasets.

\subsubsection{Initialization of the image encoder } In general, few-shot methods rely on backbone networks (e.g., ResNet \cite{ResNet_2016}) as image encoders that have been pretrained on the ImageNet data set \cite{ImageNet_2009}, which is not only very large but also entails recognition of natural objects that are associated with the segmentation of unseen target classes (e.g., bicycles, cats). However, we cannot access such a vast public dataset in seismic image interpretation. Potential reasons are the challenges associated with seismic imaging, annotation of seismic images requiring specially skilled labor (unlike natural images), and the cost involved in the process \cite{FSSS_Facies,SAM_Salt_Bodies_1,SAM_Salt_Bodies_2}. We have observed that the initialization strategy significantly affects the performance of the FSSS methods developed for natural images, such as the DGPNet \cite{dgpnet_2022}. Thus, having a pretrained backbone network for seismic datasets is important.

In the recent past, methods have been developed for the interpretation of geological features, such as the faults \cite{Faults_Self-Supervised}, salt bodies \cite{Salts_Self-Supervised}, and facies \cite{Facies_Self-Supervised} in seismic images, where self-supervised methods are used to initialize the parameters of the segmentation networks. Self-supervised contrastive methods, such as the SimCLR \cite{SimCLR_2020} and Barlow Twins \cite{BarlowTwins_2021}, are used to extract salient geological features from the \emph{unlabeled} seismic data. Barlow Twins \cite{BarlowTwins_2021} self-supervised method is used in \cite{Salts_Self-Supervised} to initialize the encoder of the segmentation network developed for the detection of salt bodies. The technique in \cite{Facies_Self-Supervised} uses reconstruction-based self-supervised learning to initialize all the parameters of the segmentation network. Both methods \cite{Salts_Self-Supervised, Facies_Self-Supervised} fine-tune the model parameters with fewer annotated examples from the target dataset to achieve comparable performance of the supervised segmentation methods. Representations learned in an unsupervised framework using deep latent variable models \cite{kingma2014auto,gens_saha_2022,avae_saha_2023,ard_vae_saha_2025,dis_avae_saha_2025} is another potential technique for initializing the network parameters.

\begin{figure*}[bh]
\centering
\includegraphics[width=0.80\textwidth]{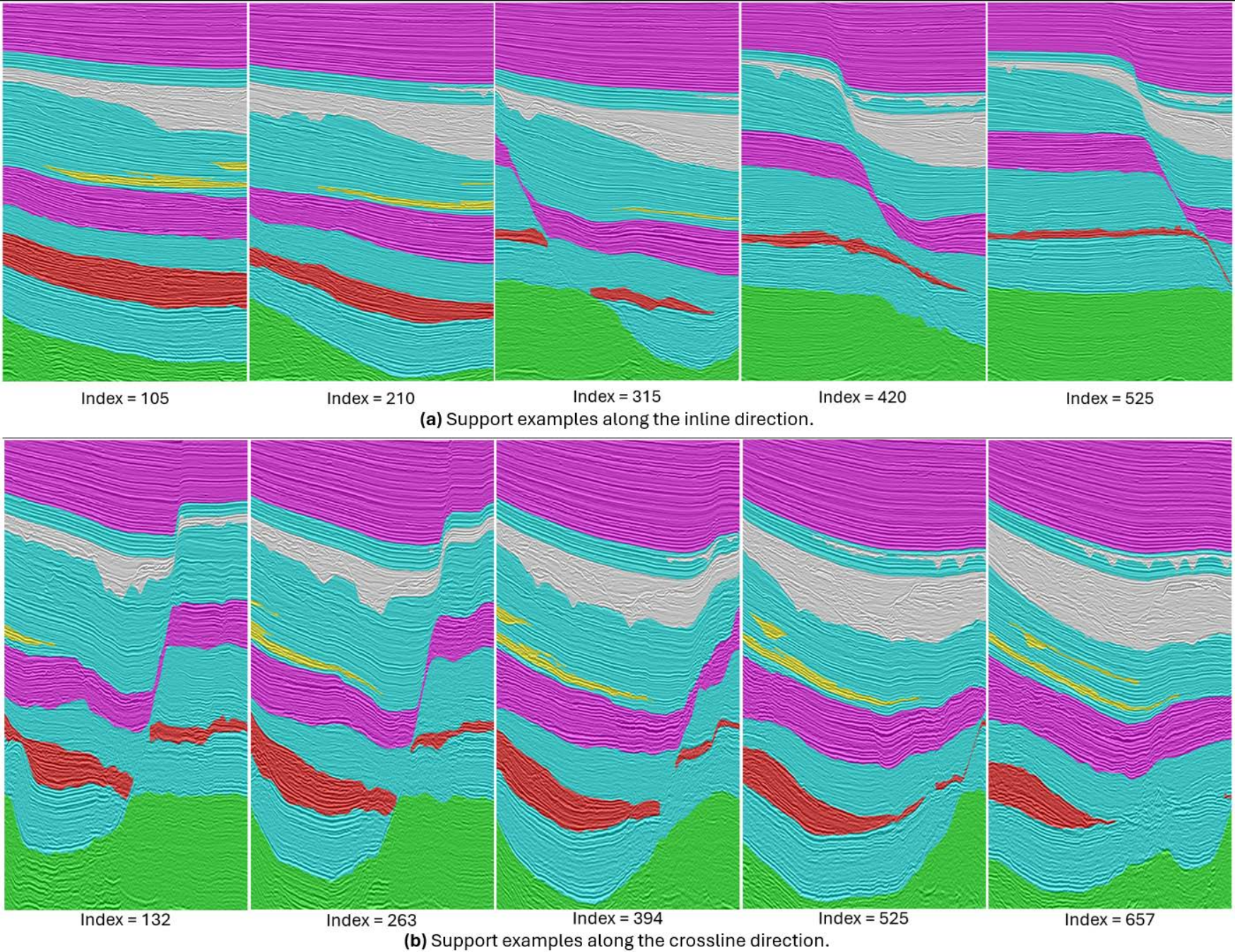}
\caption{The support set of the Parihaka dataset \cite{Parihaka_facies_new_zealand_2020} that spans through the entire volume both along the inline (slice indices=$\{105, 210, 315, 420, 525\}$) and crossline (slice indices=$\{132, 263, 394, 525, 657\}$) directions. For clarity, we overlay the input with the ground truth annotations. We observe a lot of structural variability across slices in both directions. Thus, we use a slice closest to the query image to predict facies.} \label{fig5:Parihaka_Variability}
\end{figure*}

We use a \emph{self-supervised} learning technique to initialize \emph{only} the image encoder of the AdaSemSeg with the statistics of the seismic image datasets. The advantage of self-supervised algorithms is that we \emph{do not need labeling of the data}, unlike the ImageNet \cite{ImageNet_2009}. Thus, this choice is very well aligned with the scenario where we have limited annotated training data, such as the problem of few-shot segmentation. In this work, we use the SimCLR \cite{SimCLR_2020}, a contrastive self-supervised algorithm for learning representations to assist the segmentation of seismic facies. We use all three seismic datasets studied in this work \emph{without the annotations for the facies} to train the image encoder parameters ($\theta$) of the AdaSemSeg using the SimCLR method. Under this setting, the image encoder parameters capture the statistics of the unknown target dataset. However, the mask encoder ($\psi$) and decoder ($\theta$) parameters of the AdaSemSeg that predict the facies in an input image are never exposed to any annotated samples in the target dataset. Therefore, the AdaSemSeg never learns the annotations in the target dataset. The details of the experimental setup for training the SimCLR, such as the augmentations used in generating positive and negative pairs, batch size, and other optimization configurations, are discussed in Section \ref{appsec:SimCLRTrain}.

\subsubsection{Baseline and other competing methods}\label{subsubsec:balselines} The AdaSemSeg is evaluated on the target dataset \emph{without refinement} of the network parameters on the target dataset. This evaluation setup is consistent with other FSSS methods that predict the segmentation masks on unseen target classes. We choose the AdaSemSeg trained \emph{only} on the target dataset as one of the baselines, \emph{Baseline-$1$}, and a U-Net-based \cite{UNet_2015} segmentation network, named the \emph{Baseline-$2$}, as another baseline, which is also trained on the target dataset. Besides, we compare the performance with a prototype-based \cite{Proto_2017} FSSS method for seismic facies \cite{facies_protoseg_2023}, referred to herein as the \emph{ProtoSemSeg}. 

\emph{Transfer learning} has been widely used in deep learning research, where the last few layers of a deep neural network are retrained (sometimes with modifications) using a few examples from the target dataset. It is used to interpret seismic images, such as classifying seismic facies \cite{facies_transferlearn_2019} and detecting faults \cite{f3_faultlines_transferlearn_2019}. We train the Baseline-$2$ using transfer learning, where the model is first trained on patches from the source data, e.g., the F3 and Penobscot volume, and then we fine-tune the parameters on a handful of annotated samples in the target class, i.e.,  the Parihaka dataset with respect to this example. The last layer of the decoder in the Baseline-$2$ is customized to the number of classes in the facies dataset, such as $7$ for the Penobscot and $6$ in the case of the Parihaka and F3 datasets.

\subsubsection{Evaluation metrics}\label{subsubsec:metrics} The performance of the AdaSemSeg and other methods is evaluated under different metrics used in the literature \cite{f3_facies_netherlands_2019,seismic_facies_2022}, such as the pixel accuracy (PA), class accuracy, mean class accuracy (MCA), intersection over union (IoU), and $F_1$ score. The IoU and $F_1$ scores are weighted by the frequency of the classes, denoted by the FwIoU and Fw$F_1$ scores, respectively. We use whole slices to evaluate all the methods (supported by convolutional neural networks) for practical use cases and to avoid the complexity of patch stitching \cite{seismic_facies_2022}. Moreover, patch stitching offers no additional performance benefits relative to whole slices.

\begin{table*}[th]
\centering
\caption{Evaluation of the AdaSemSeg on the F3, Penobscot, and Parihaka datasets using $K=5$ support examples or the \emph{nearest slice} in the support set. The use of the $K=5$ support examples for evaluating the query image is indicated by \xmark \:, and the \checkmark \: represents the use of the \emph{nearest slice} as the support example. The best performance of the AdaSemSeg under different evaluation scenarios is highlighted in \textbf{bold}.}\label{tab1:AdaSemSeg_evaluation_nearest_slice}
\resizebox{\textwidth}{!}
{
\begin{tabular}{| c | c c | c c c c c c c c | c c | c c c c c c c c |}
\hline
\multirow{2}{*}{\parbox{1.5cm}{\centering Dataset}} && \multirow{2}{*}{\parbox{1.5cm}{\centering Nearest slice}} & \multicolumn{8}{c|}{inline}  && \multirow{2}{*}{\parbox{1.5cm}{\centering Nearest slice}} & \multicolumn{8}{c|}{crossline} \\ \cline{4-11} \cline{14-21}
&& &&PA &&MCA &&FwIoU &&Fw$F_1$ && &&PA &&MCA &&FwIoU &&Fw$F_1$\\ \hline
\multirow{2}{*}{F3} &&\xmark &&$\mathbf{0.89}$ &&$\mathbf{0.79}$ &&$\mathbf{0.81}$ &&$\mathbf{0.89}$  &&\xmark &&$\mathbf{0.87}$ &&$\mathbf{0.73}$ &&$\mathbf{0.80}$ &&$\mathbf{0.88}$\\
&&\checkmark &&0.85 &&0.73 &&0.78 &&0.85 &&\checkmark &&0.80 &&0.58 &&0.71 &&0.81\\ \hline
\multirow{2}{*}{Penobscot} &&\xmark &&0.95 &&$\mathbf{0.95}$ &&0.91 &&0.96  &&\xmark &&$\mathbf{0.97}$ &&$\mathbf{0.95}$ &&$\mathbf{0.93}$ &&$\mathbf{0.96}$\\
&&\checkmark &&$\mathbf{0.96}$ &&$\mathbf{0.95}$ &&$\mathbf{0.94}$ &&$\mathbf{0.97}$ &&\checkmark &&0.96 &&0.94 &&0.92 &&0.95\\ \hline
\multirow{2}{*}{Parihaka} &&\xmark &&0.78 &&0.68 &&0.66 &&0.79 &&\xmark &&0.79 &&0.65 &&0.67 &&0.80\\
&&\checkmark &&$\mathbf{0.86}$ &&$\mathbf{0.76}$ &&$\mathbf{0.76}$ &&$\mathbf{0.86}$  &&\checkmark &&$\mathbf{0.84}$ &&$\mathbf{0.68}$ &&$\mathbf{0.74}$ &&$\mathbf{0.85}$\\ \hline
\end{tabular}
}
\end{table*}

\begin{table*}[t]
\centering
\caption{Comparison of the AdaSemSeg (w/o refinement) with baselines trained \emph{only} on the target datasets. NA under several class indices for different datasets indicates the absence of the corresponding class indices in the ground truth annotations. The performance of the best method is highlighted in \textbf{bold}.}\label{tab2:AdaSemSeg_baselines}
\resizebox{\textwidth}{!}
{
\begin{tabular}{| c | c | c | c c c c c c c c c c c c c c c | c | c | c |}
\hline
\multirow{2}{*}{\parbox{1.5cm}{\centering Target Dataset}} & \multirow{2}{*}{Method} & \multirow{2}{*}{PA} && \multicolumn{13}{c}{Class accuracy} && \multirow{2}{*}{MCA} & \multirow{2}{*}{FwIoU} & \multirow{2}{*}{Fw$F_1$}\\ \cline{5-17}
&&&&1 && 2 && 3 && 4 && 5 && 6 && 7&&&&\\ \hline
\multirow{3}{*}{\parbox{1.5cm}{\centering Parihaka inline}} & AdaSemSeg & 0.86 && 0.98 && 0.87 && 0.37 && 0.80 && NA && 0.76 && NA && 0.76 & 0.76 & 0.86\\
& Baseline$-1$ & $\mathbf{0.91}$ && $\mathbf{0.99}$ && $\mathbf{0.92}$ && 0.75 && 0.83 && NA && $\mathbf{0.87}$ && NA && $\mathbf{0.87}$ & $\mathbf{0.83}$ & $\mathbf{0.90}$\\
& Baseline$-2$ & 0.87 && 0.85 && 0.87 && $\mathbf{0.76}$ && $\mathbf{0.88}$ && NA && $\mathbf{0.87}$ && NA && 0.85 & 0.77 & 0.86 \\ [1.5ex] \hline 
\multirow{3}{*}{\parbox{1.5cm}{\centering Parihaka crossline}} & AdaSemSeg & 0.84 && $\mathbf{0.84}$ && $\mathbf{0.91}$ && 0.63 && 0.82 && 0.02 && 0.83 && NA && 0.68 & 0.74 & 0.85\\
& Baseline$-1$ & $\mathbf{0.91}$ && 0.83 && $\mathbf{0.91}$ && 0.89 && 0.90 && $\mathbf{0.82}$ && $\mathbf{0.97}$ && NA && $\mathbf{0.89}$ & $\mathbf{0.85}$ & $\mathbf{0.92}$\\
& Baseline$-2$ & 0.86 && 0.76 && 0.88 && $\mathbf{0.92}$ && $\mathbf{0.93}$ && 0.65 && 0.77 && NA && 0.82 & 0.78 & 0.88 \\ [1.5ex] \hline
\multirow{3}{*}{\parbox{1.5cm}{\centering Penobscot inline}} & AdaSemSeg & 0.95 && 0.97 && 0.99 && 0.92 && 0.94 && 0.99 && 0.90 && 0.91 && 0.95 & 0.91 & 0.96\\
& Baseline$-1$ & $\mathbf{0.97}$ && $\mathbf{0.99}$ && $\mathbf{1.00}$ && $\mathbf{0.99}$ && 0.98 && 0.98 && $\mathbf{1.00}$ && $\mathbf{0.92}$ && $\mathbf{0.98}$ & $\mathbf{0.95}$ & $\mathbf{0.98}$\\
& Baseline$-2$ & $\mathbf{0.97}$ && $\mathbf{0.99}$ && 0.99 && $\mathbf{0.99}$ && $\mathbf{0.99}$ && $\mathbf{1.00}$ && 0.99 && 0.90 && $\mathbf{0.98}$ & $\mathbf{0.95}$ & $\mathbf{0.98}$ \\ [1.5ex] \hline
\multirow{3}{*}{\parbox{1.5cm}{\centering Penobscot crossline}} & AdaSemSeg & 0.97 && 0.96 && 0.97 && 0.92 && 0.88 && 0.98 && 0.96 && 0.98 && 0.95 & 0.93 & 0.96\\
& Baseline$-1$ & 0.97 && $\mathbf{0.98}$ && $\mathbf{0.98}$ && $\mathbf{0.99}$ && 0.94 && 0.96 && $\mathbf{0.99}$ && $\mathbf{0.97}$ && 0.97 & 0.94 & $\mathbf{0.97}$\\
& Baseline$-2$ & $\mathbf{0.98}$ && $\mathbf{0.98}$ && 0.97 && 0.98 && $\mathbf{0.97}$ && $\mathbf{0.99}$ && $\mathbf{0.99}$ && $\mathbf{0.97}$ && $\mathbf{0.98}$ & $\mathbf{0.95}$ & $\mathbf{0.97}$ \\ [1.5ex] \hline
\multirow{3}{*}{\parbox{1.5cm}{\centering F3 inline}} & AdaSemSeg & 0.89 &&0.97 && $\mathbf{0.94}$ && 0.96 && 0.76 && 0.66 && 0.43 && NA && 0.79 & 0.81 & 0.89\\
& Baseline$-1$ & $\mathbf{0.91}$ && $\mathbf{0.99}$ && $\mathbf{0.94}$ && $\mathbf{0.97}$ && 0.81 && $\mathbf{0.73}$ && $\mathbf{0.88}$ && NA && $\mathbf{0.89}$ & $\mathbf{0.86}$ & $\mathbf{0.91}$\\
& Baseline$-2$ & 0.88 && $\mathbf{0.99}$ && $\mathbf{0.94}$ && 0.96 && $\mathbf{0.89}$ && 0.56 && 0.67 && NA && 0.84 & 0.79 & 0.87 \\ [1.5ex] \hline
\multirow{3}{*}{\parbox{1.5cm}{\centering F3 crossline}} & AdaSemSeg & 0.87 && 0.94 && $\mathbf{0.89}$ && $\mathbf{0.88}$ && 0.68 && 0.18 && 0.80 && NA && 0.73 & 0.80 & 0.88\\
& Baseline$-1$ & 0.86 && $\mathbf{0.98}$ && 0.86 && 0.85 && 0.61 && 0.17 && $\mathbf{0.98}$ && NA && 0.74 & 0.80 & 0.87\\
& Baseline$-2$ & $\mathbf{0.89}$ && $\mathbf{0.98}$ && 0.85 && 0.87 && $\mathbf{0.82}$ && $\mathbf{0.36}$ && 0.88 && NA && $\mathbf{0.79}$ & $\mathbf{0.82}$ & $\mathbf{0.89}$ \\ [1.5ex] \hline
\end{tabular}
}
\end{table*}

\begin{figure*}[th]
\centering
\includegraphics[width=0.79\textwidth]{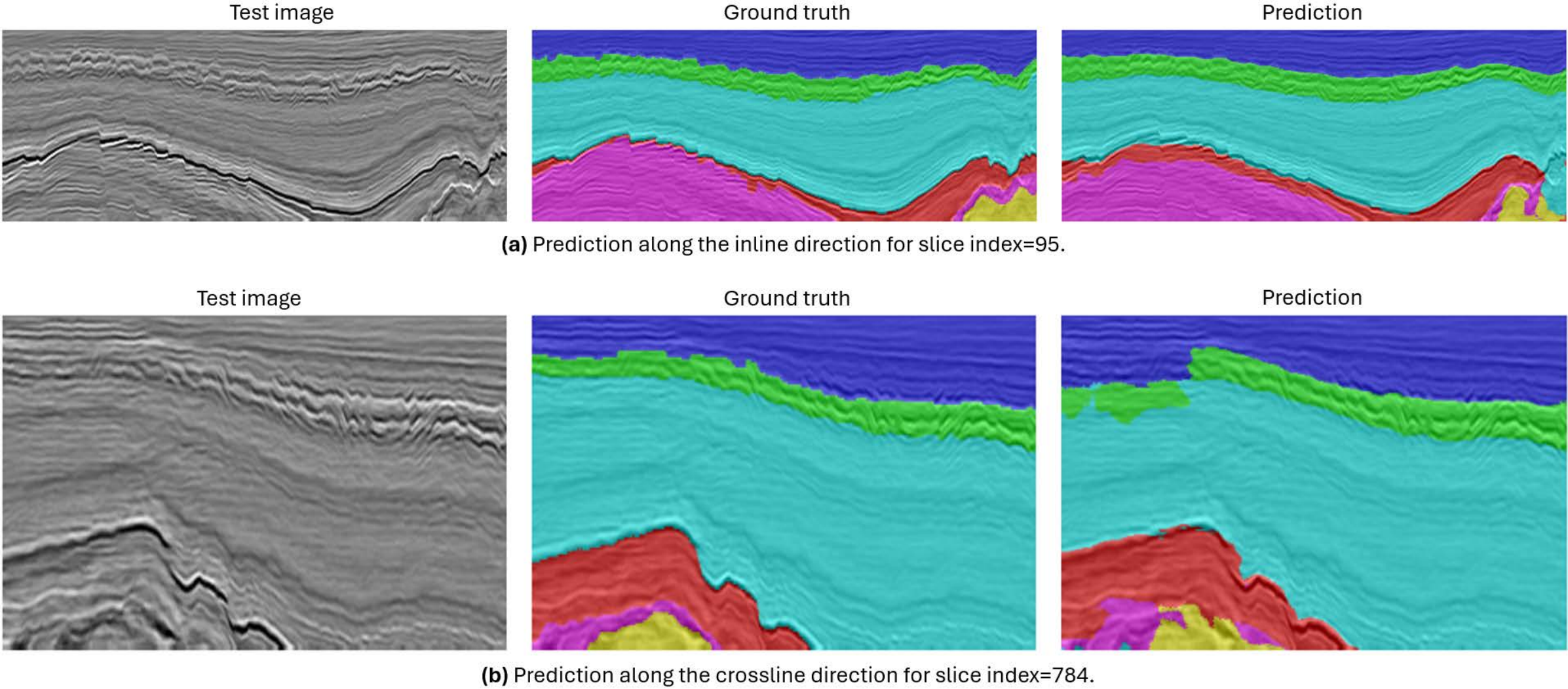}
\caption{Prediction of facies on the test data of the F3 dataset \cite{f3_facies_netherlands_2019} along the inline and crossline directions by the AdaSemSeg trained in the $5-$shot setup on the Parihaka and Penobscot datasets. The AdaSemSeg uses only $5-$ support examples (shown in Fig. \ref{fig4:F3_Support_Set}) to predict the facies on the unseen dataset.} \label{fig4:f3_facies_predictions}
\end{figure*}

\begin{figure*}[t]
\centering
\includegraphics[width=0.79\textwidth]{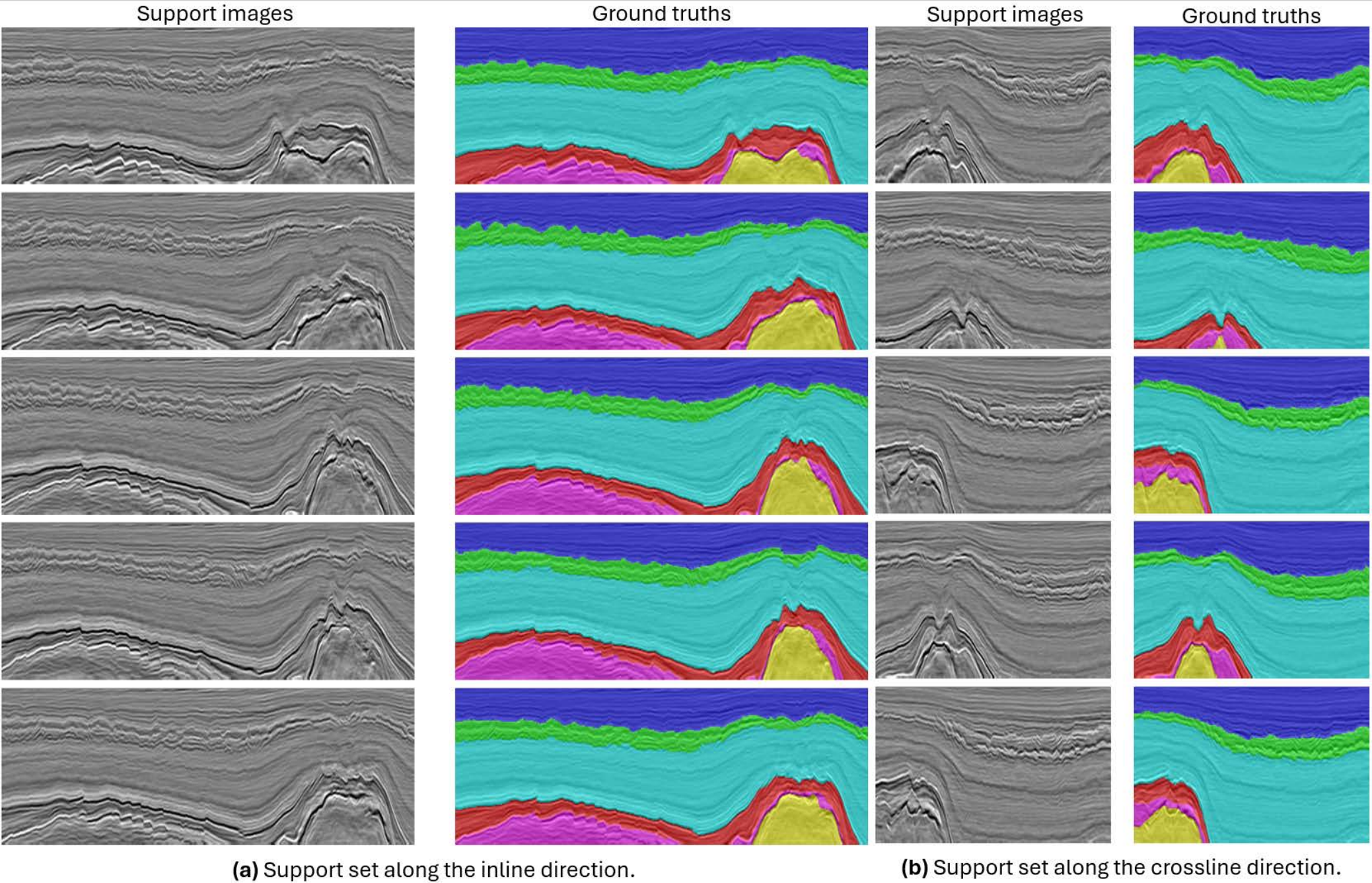}
\caption{The support set of the F3 facies data \cite{f3_facies_netherlands_2019} that spans through the entire volume both along the inline (slice indices=$\{15, 111, 207, 303, 400\}$) and crossline (slice indices=$\{0, 171, 343, 514, 686\}$) directions. The AdaSemSeg uses the support examples to predict facies on the test slices of the F3, as shown in Fig. \ref{fig4:f3_facies_predictions}.
} \label{fig4:F3_Support_Set}
\end{figure*}

\subsection{Experimental Results}
For all the results discussed in this section, we assume that after completing the meta-training on source datasets, the AdaSemSeg is evaluated on the target dataset \emph{without refinement} of the network parameters on samples from the target dataset. The details of the image encoder (IE) and decoder (D) used in the AdaSemSeg, Baseline-$2$, and ProSemSeg, along with the mask encoder (ME) used by the AdaSemSeg are reported in Appendix \ref{appsec:NN_arch}. Refer to Appendix \ref{appsec:Train_spec} for the configuration of the optimizer, learning rate scheduler, augmentations, and other experimental details.

\subsubsection{The use of the support examples for evaluation} The AdaSemSeg uses $K=\{1, 5\}$ examples from the training volume of the target dataset as the support set ($S$) that spans the whole volume to predict the facies on the query images, $I^q$s, which are samples in the test set of the target dataset. For example, the indices of the support set along the crossline direction for the F3 facies dataset 
are $S_i=\{0, 171, 343, 514, 686\}$ and the test data starts at the index $700$ (refer to Fig. \ref{fig3:Train_val_test_split} for details). Instead of $K$ support images, we can use a single sample in the support set closest to the query slice as the support sample, referred to herein as the \emph{nearest slice}. Under this evaluation scenario, we use the slice at index $686 \in S_i$ as the support set (S) to predict the mask for the same test data. We hypothesize that the second approach to constructing the support set is likely more effective when significant morphological changes are present along an axis, as shown in Fig. \ref{fig5:Parihaka_Variability} for the Parihaka dataset.

We evaluate the AdaSemSeg using both the sampling techniques of the support set (discussed in the previous paragraph) on all the datasets, separately along the inline and crossline directions of the test data. From the results reported in Table \ref{tab1:AdaSemSeg_evaluation_nearest_slice}, we observe that using more support examples is favorable for the F3 dataset along both directions. We do not observe many variations in performance between the sampling techniques of the support set for the Penobscot dataset. Thus, we use all the $K$ examples in the support set for evaluation on the F3 and Penobscot datasets. However, the performance on the Parihaka dataset is more effective using the nearest slice in the support set along both directions due to the structural variations in the horizons. Therefore, we use the nearest slice to evaluate the Parihaka dataset for subsequent analyses.

\begin{table*}[th]
\centering
\caption{Comparison of the AdaSemSeg with ProtoSemSeg and transfer learning under the $1$-shot and $5$-shot setup. The best performance is highlighted in \textbf{bold}.}\label{tab3:AdaSemSeg_Few_shot}
\resizebox{\textwidth}{!}
{
\begin{tabular}{| c | c | c c c c c c | c c c c c c |}
\hline
\multirow{2}{*}{\parbox{1.5cm}{\centering Target dataset}} & \multirow{2}{*}{Metric} && \multicolumn{5}{c|}{$1-$shot} && \multicolumn{5}{c|}{$5-$shot} \\ \cline{3-8} \cline{9-14}
&&&AdaSemSeg && ProtoSemSeg && \parbox{1.5cm}{\centering Transfer learning} && AdaSemSeg && ProtoSemSeg && \parbox{1.5cm}{\centering Transfer learning}\\ \hline
\multirow{4}{*}{\parbox{1.5cm}{\centering Parihaka inline}}&PA&&$\mathbf{0.84}$ && 0.51 && 0.49 && $\mathbf{0.86}$ && 0.56 && 0.59\\
&MCA&&$\mathbf{0.75}$ && 0.45 && 0.60 && $\mathbf{0.76}$ && 0.53 && 0.66\\
&FwIoU&&$\mathbf{0.74}$ && 0.36 && 0.37 && $\mathbf{0.76}$ && 0.43 && 0.45\\
&Fw$F_1$&&$\mathbf{0.84}$ && 0.52 && 0.54 && $\mathbf{0.86}$ && 0.58 && 0.62\\ [1.5ex] \hline
\multirow{4}{*}{\parbox{1.5cm}{\centering Parihaka crossline}}&PA&&$\mathbf{0.82}$ && 0.60 && 0.55 && $\mathbf{0.84}$ && 0.61 && 0.72\\
&MCA&&$\mathbf{0.71}$ && 0.42 && 0.66 && $\mathbf{0.68}$ && 0.58 && 0.74\\
&FwIoU&&$\mathbf{0.71}$ && 0.46 && 0.41 && $\mathbf{0.74}$ && 0.48 && 0.59\\
&Fw$F_1$&&$\mathbf{0.83}$ && 0.62 && 0.58 && $\mathbf{0.85}$ && 0.64 && 0.74\\ [1.5ex] \hline
\multirow{4}{*}{\parbox{1.5cm}{\centering Penobscot inline}}&PA&&$\mathbf{0.92}$ && 0.58 && 0.67 && $\mathbf{0.95}$ && 0.71 && 0.89\\
&MCA&&$\mathbf{0.92}$ && 0.45 && 0.75 && $\mathbf{0.95}$ && 0.62 && 0.93\\
&FwIoU&&$\mathbf{0.85}$ && 0.42 && 0.53 && $\mathbf{0.91}$ && 0.56 && 0.81\\
&Fw$F_1$&&$\mathbf{0.93}$ && 0.58 && 0.67 && $\mathbf{0.96}$ && 0.71 && 0.89\\ [1.5ex] \hline
\multirow{4}{*}{\parbox{1.5cm}{\centering Penobscot crossline}}&PA&&$\mathbf{0.91}$ && 0.54 && 0.62 && $\mathbf{0.97}$ && 0.70 && 0.87\\
&MCA&&$\mathbf{0.90}$ && 0.40 && 0.70 && $\mathbf{0.95}$ && 0.56 && 0.91\\
&FwIoU&&$\mathbf{0.83}$ && 0.38 && 0.47 && $\mathbf{0.93}$ && 0.56 && 0.76\\
&Fw$F_1$&&$\mathbf{0.90}$ && 0.54 && 0.62 && $\mathbf{0.96}$ && 0.70 && 0.86\\ [1.5ex] \hline
\multirow{4}{*}{\parbox{1.5cm}{\centering F3 inline}}&PA&&$\mathbf{0.85}$ && 0.57 && 0.83 && $\mathbf{0.89}$ && 0.69 && 0.85\\
&MCA&&0.72 && 0.45 && $\mathbf{0.80}$ && $\mathbf{0.79}$ && 0.56 && $\mathbf{0.79}$\\
&FwIoU&&$\mathbf{0.77}$ && 0.40 && 0.73 && $\mathbf{0.81}$ && 0.53 && 0.75\\
&Fw$F_1$&&$\mathbf{0.85}$ && 0.55 && 0.84 && $\mathbf{0.89}$ && 0.68 && 0.84\\ [1.5ex] \hline
\multirow{4}{*}{\parbox{1.5cm}{\centering F3 crossline}}&PA&&$\mathbf{0.87}$ && 0.64 && 0.75 && $\mathbf{0.87}$ && 0.77 && 0.81\\
&MCA&&$\mathbf{0.68}$ && 0.36 && 0.62 && $\mathbf{0.73}$ && 0.44 && 0.70\\
&FwIoU&&$\mathbf{0.79}$ && 0.50 && 0.68 && $\mathbf{0.80}$ && 0.65 && 0.72\\
&Fw$F_1$&&$\mathbf{0.87}$ && 0.64 && 0.79 && $\mathbf{0.88}$ && 0.77 && 0.82\\ [1.5ex] \hline
\end{tabular}
}
\end{table*}

\begin{table*}[t]
\centering
\caption{Effect of the initialization strategy on the performance of the AdaSemSeg when evaluated on the Parihaka datasets using $k=\{1, 5\}$ support examples. The performance of the best method is presented in \textbf{bold}.}\label{tab5:AdaSemSeg_parihaka_initialiazation}
\resizebox{\textwidth}{!}
{
\begin{tabular}{|c | c | c c c c c c c c | c c c c c c c c |}
\hline
\multirow{2}{*}{\parbox{1.5cm}{\centering Shots}} & \multirow{2}{*}{\parbox{3.5cm}{\centering Initialization of the image encoder}} &\multicolumn{8}{c|}{inline}& \multicolumn{8}{c|}{crossline} \\ \cline{3-10} \cline{11-18}
&&&PA &&MCA &&FwIoU &&Fw$F_1$ &&PA &&MCA &&FwIoU &&Fw$F_1$\\ \hline
\multirow{2}{*}{\parbox{1.5cm}{\centering 1}} &Random && 0.61 &&0.58 &&0.48 &&0.64 && 0.56 &&0.50 &&0.42 &&0.59 \\
&SimCLR &&$\mathbf{0.84}$ &&$\mathbf{0.75}$ &&$\mathbf{0.74}$ &&$\mathbf{0.84}$ && $\mathbf{0.82}$ &&$\mathbf{0.71}$ &&$\mathbf{0.71}$ &&$\mathbf{0.83}$ \\[1.5 ex] \hline
\multirow{2}{*}{\parbox{1.5cm}{\centering 5}} &Random && 0.72 &&0.66 &&0.59 &&0.72 && 0.55 &&0.57 &&0.43 &&0.60 \\
&SimCLR && $\mathbf{0.86}$ &&$\mathbf{0.76}$ &&$\mathbf{0.76}$ &&$\mathbf{0.86}$ && $\mathbf{0.84}$ &&$\mathbf{0.68}$ &&$\mathbf{0.74}$ &&$\mathbf{0.85}$ \\[1.5 ex] \hline
\end{tabular}
}
\end{table*}

\subsubsection{Comparison with baseline methods trained on the target dataset}\label{subsubsec:baselines} We evaluate the performance of the AdaSemSeg and the baselines, Baseline-$1$ and Baseline-$2$, discussed in section \ref{subsubsec:balselines} for all the datasets. Both baselines, \emph{Baseline-$1$} and \emph{Baseline-$2$}, are trained on samples from the target dataset, unlike the AdaSemSeg. From the results reported in Table \ref{tab2:AdaSemSeg_baselines}, we observe that the Baseline-$1$, i.e., the AdaSemSeg trained only on the target dataset, produces the best performance under most of the evaluation scenarios across all the datasets. This is possibly due to the flexible Gaussian process regression in the latent space of the Baseline-$1$. However, for the F3 dataset along the crossline direction, the Baseline$-2$ outperforms all the methods under different evaluation metrics. 

Though the parameters of the AdaSemSeg are not tuned to interpret the images from the target dataset, its performance is comparable to the baselines trained on the target datasets. Moreover, we observe a marginal difference in the performance of the AdaSemSeg relative to the baselines on the Penobscot and F3 datasets. This demonstrates the generalization capability of few-shot learning, where the statistics of the support set (few annotated examples, $K=\{1, 5\}$) are effectively used to make reasonable predictions on unseen query images. However, the difference with the baselines is noticeable in the Parihaka dataset, possibly due to the complexity of the dataset as indicated by the visual interpretation of geological features relative to the Penobscot and F3 datasets. 

The experimental results show that the seismic features learned from the Penobscot and F3 datasets (source datasets) in the meta-training stage do not generalize well to the Parihaka dataset (target dataset). However, the complex Parihaka dataset in the meta-training stage is helping AdaSemSeg to produce impressive predictions on unseen samples from the F3 and Penobscot datasets. Overall, this experiment demonstrates the strength of AdaSemSeg in adapting to the unseen target dataset. Fig. \ref{fig4:f3_facies_predictions} illustrates the effectiveness of the AdaSemSeg in predicting facies on the unseen F3 dataset (target dataset) using the features learned from the Parihaka and Penobscot datasets (source datasets). The AdaSemSeg uses $5$ support examples from the F3 facies dataset along the inline and crossline as shown in Fig. \ref{fig4:F3_Support_Set} directions to predict the segmentation masks shown in Fig. \ref{fig4:f3_facies_predictions}. We must remember that the support examples from the F3 dataset (i.e., the target dataset) are not used to fine-tune the parameters of the AdaSemSeg. Predictions on other datasets and the corresponding support examples used for predicting facies are shown in Appendix \ref{appsec:Exp_res}.

\subsubsection{Comparison with other competing methods}\label{subsubsec:competing_methods}
In this experiment, we compare the performance of the AdaSemSeg with ProtoSemSeg and transfer learning under $1-$shot and $5-$shot scenarios, outlined in Table \ref{tab3:AdaSemSeg_Few_shot}. For the transfer learning, we first train the Baseline-$2$ with patches extracted from the source data, e.g., the F3 and Penobscot volume, and fine-tuned the trained parameters on a handful of annotated data from the target data, i.e., the Parihaka volume regarding this example. We use patches extracted from $\{1, 5\}$ slice(s) in the target dataset's support set to fine-tune the Baseline-$2$ parameters. However, the AdaSemSeg and ProtoSemSeg are \emph{not} trained on the target dataset.

The AdaSemSeg outperforms all the competing methods under multiple evaluation metrics. This experiment demonstrates the strength of the GP-based few-shot learning in the AdaSemSeg over the prototype-based few-shot segmentation method. Moreover, fine-tuning the segmentation network on the target dataset in transfer learning did not help in learning generalized representations that would assist in identifying facies in the target dataset.

\begin{table*}[t]
\centering
\caption{Effect of the data augmentation strategies on the performance of the AdaSemSeg when evaluated on the Parihaka datasets using only $k=\{1\}$ support example. The performance of the best augmentation method is presented in \textbf{bold}. \textbf{All} indicates the use of all data augmentation methods in training the AdaSemSeg.}\label{tab6:AdaSemSeg_data_augmentations}
\resizebox{\textwidth}{!}
{
\begin{tabular}{| c | c c c c c c c c | c c c c c c c c |}
\hline
\multirow{2}{*}{\parbox{3.5cm}{\centering Augmentation method}} &\multicolumn{8}{c|}{inline}& \multicolumn{8}{c|}{crossline} \\ \cline{2-9} \cline{10-17}
&&PA &&MCA &&FwIoU &&Fw$F_1$ &&PA &&MCA &&FwIoU &&Fw$F_1$\\ \hline
None &&0.83 &&0.83 &&0.73 &&0.84 &&0.76 &&0.73 &&0.66 &&0.79 \\ [1.0 ex] \hline
RandomRotate &&$\mathbf{0.86}$ &&0.76 &&$\mathbf{0.76}$ &&$\mathbf{0.85}$ &&$\mathbf{0.82}$ &&0.68 &&$\mathbf{0.72}$ &&$\mathbf{0.83}$ \\ [1.0 ex] \hline
RandomHorizontalFlip &&0.85 &&0.79 &&0.74 &&0.84 &&0.81 &&0.67 &&0.69 &&0.82 \\ [1.0 ex] \hline
GaussianBlur &&0.83 &&$\mathbf{0.84}$ &&0.73 &&0.84 &&0.75 &&$\mathbf{0.76}$ &&0.65 &&0.79 \\ [1.0 ex] \hline
GaussNoise && 0.83 &&0.81 &&0.73 &&0.83 &&0.77 &&0.71 &&0.67 &&0.80 \\ [1.0 ex] \hline
All && 0.84 &&0.75 &&0.74 &&0.84 &&$\mathbf{0.82}$ &&0.71 &&0.71 &&$\mathbf{0.83}$ \\ [1.0 ex] \hline
\end{tabular}
}
\end{table*}

\subsubsection{Ablation study on the initialization of the image encoder} In another experiment, we study the effect of the image encoder's initialization on the AdaSemSeg's performance. In this experiment, we initialize the image encoder randomly and compare its performance with an image encoder trained using the SimCLR contrastive learning algorithm. We evaluate the performance of the proposed method on the Parihaka dataset under $1$ and $5$-shot scenarios, and the metric scores reported in Table \ref{tab5:AdaSemSeg_parihaka_initialiazation} explain the importance of the initialization of the image encoder in the AdaSemSeg with the statistics of the seismic datasets. This observation is consistent with the performance of other FSSS methods developed for natural images, such as the initialization of the image encoder of the DGPNet \cite{dgpnet_2022} with the statistics of the ImageNet dataset \cite{ImageNet_2009}.

\subsubsection{Ablation study on data augmentation techniques} We use the following data augmentation techniques in training the AdaSemSeg: RandomRotate, RandomHorizontalFlip, GaussianBlur, and GaussNoise. To study the impact of an individual augmentation strategy on the performance of the AdaSemSeg, we train the AdaSemSeg separately using a single data augmentation technique. We follow all the details reported in Appendix \ref{appsec:Train_spec} for training the AdaSemSeg on the source datasets. Each trained model is evaluated on the same target dataset. In another scenario, we disregard all data augmentation strategies during the training of AdaSemSeg. To evaluate the dependence of the AdaSemSeg on data augmentation, we perform our analysis on the challenging Parihaka dataset using only $k=\{1\}$ support example. In this analysis, we train the AdaSemSeg on the Penobscot and F3 datasets and evaluate the trained model on the \textit{unseen} Parihaka dataset. The comparative analysis of the effects of different augmentation techniques is reported in Table \ref{tab6:AdaSemSeg_data_augmentations}.

Out of all the augmentation methods studied in this work, we observe that the \textit{RandomRotate} data augmentation method influences the performance of the AdaSemSeg under multiple evaluation metrics, except for the mean classification accuracy (MCA). Along the inline direction of the Parihaka dataset, the AdaSemSeg is \textit{unaffected} by the use of any data augmentation technique, which demonstrates the robustness of the proposed method. However, we observe slight improvements along the crossline direction with the use of data augmentation for all the evaluation metrics. To take advantage of the collective benefits of different augmentation techniques, we use all four augmentation methods in training the AdaSemSeg along with the baselines and competing methods.

\subsubsection{Ablation study on the model architecture} The Baseline-2 model studied in this work is a regular image segmentation network that uses an image encoder and image decoder as used in the UNet \cite{UNet_2015} architecture for predicting the seismic facies. In comparison to the AdaSemSeg, we do not use a mask encoder in the Baseline-2 model and do not perform Gaussian process regression in the latent spaces. Therefore, the Baseline-2 model is an ablation study on the architecture of the proposed AdaSemSeg. The Baseline-2 model is used in two different scenarios. In the first scenario, we train the Baseline-2 model only on patches extracted from the 2D slices in the support set of the target dataset and evaluate the trained model on the test slices of the target dataset. The performance of the Baseline-2 used in this context is reported in Table \ref{tab2:AdaSemSeg_baselines} and we discuss the results in Section \ref{subsubsec:baselines}. In the second scenario, we use the Baseline-2 model for transfer learning, where the model is first trained on patches of the source datasets and evaluated on the unseen target dataset. Results of this experiment are discussed in Section \ref{subsubsec:competing_methods} and metric scores are reported in Table \ref{tab3:AdaSemSeg_Few_shot}. Across both scenarios, the AdaSemSeg outperforms or is comparable to the performance of the Baseline-2 model.

\begin{table*}[t]
\centering
\caption{Average time taken (in seconds) by the competing methods using GPU and CPU under the $1$-shot setup for different target datasets. We provide the resolution of the 2D slices for all the datasets. The best performance is highlighted in \textbf{bold}. As expected, the Baseline-$2$ model takes the minimum amount of time for both GPU and CPU.}\label{tab7:AdaSemSeg_Few_shot_GPU_CPU}
\resizebox{\textwidth}{!}
{
\begin{tabular}{| c | c | c c c c c c | c c c c c c |}
\hline
\multirow{2}{*}{\parbox{1.5cm}{\centering Target dataset}} &\multirow{2}{*}{Resolution} && \multicolumn{5}{c|}{GPU (in seconds)} && \multicolumn{5}{c|}{CPU (in seconds)} \\ \cline{3-8} \cline{9-14}
&&&AdaSemSeg && ProtoSemSeg && \parbox{1.5cm}{\centering Baseline-$2$} && AdaSemSeg && ProtoSemSeg && \parbox{1.5cm}{\centering Baseline-$2$}\\ \hline
\multirow{2}{*}{\parbox{2.4cm}{\centering Parihaka inline}}
&\multirow{2}{*}{\parbox{1.5cm}{\centering $992 \times 640$}}
&&$0.52$ &&$0.35$ &&$\mathbf{0.15}$ 
&&$55.34$ &&$54.95$ &&$\mathbf{25.11}$
\\ [1.2ex] \hline
\multirow{2}{*}{\parbox{2.4cm}{\centering Parihaka crossline}}
&\multirow{2}{*}{\parbox{1.5cm}{\centering $992 \times 512$}}
&&$0.41$ &&$0.27$ &&$\mathbf{0.08}$
&&$44.82$ &&$45.13$ &&$\mathbf{20.07}$
\\ [1.5ex] \hline
\multirow{2}{*}{\parbox{2.4cm}{\centering Penobscot inline}}
&\multirow{2}{*}{\parbox{1.5cm}{\centering $864 \times 384$}}
&&$0.62$ &&$0.22$ &&$\mathbf{0.10}$
&&$35.03$ &&$31.18$ &&$\mathbf{12.89}$
\\ [1.5ex] \hline
\multirow{2}{*}{\parbox{2.4cm}{\centering Penobscot crossline}}
&\multirow{2}{*}{\parbox{1.5cm}{\centering $864 \times 384$}}
&&$0.59$ &&$0.22$ &&$\mathbf{0.10}$
&&$34.73$ &&$32.17$ &&$\mathbf{12.25}$
\\ [1.5ex] \hline
\multirow{2}{*}{\parbox{2.4cm}{\centering F3 inline}}
&\multirow{2}{*}{\parbox{1.5cm}{\centering $256 \times 672$}}
&&$0.34$ &&$0.11$ && $\mathbf{0.04}$
&&$13.84$ &&$6.95$ &&$\mathbf{6.20}$
\\ [1.5ex] \hline
\multirow{2}{*}{\parbox{2.4cm}{\centering F3 crossline}}
&\multirow{2}{*}{\parbox{1.5cm}{\centering $256 \times 384$}}
&&$0.23$ &&$0.09$ &&$\mathbf{0.03}$
&&$9.00$ &&$4.23$ &&$\mathbf{3.84}$
\\ [1.5ex] \hline
\end{tabular}
}
\end{table*}

\subsubsection{Inference time using CPU and GPU} To solve the multi-class segmentation task, the AdaSemSeg combines multiple binary predictions during training (meta-training) and evaluation (meta-testing) using a shared DGPNet. Refer to Figure \ref{fig:DGPNet_Multiclass} and \ref{fig:Inference_AdaSemSeg} for the visual illustrations of the training and inference process of the AdaSemSeg, respectively. The AdaSemSeg iterates over all the classes in the source and target datasets as shown in Algorithm \ref{alg:Meta-train-AdaSemSeg} (meta-training) and \ref{alg:Meta-test-AdaSemSeg} (meta-testing), respectively. For the deployment of the proposed method in solving real-life applications, it is important to analyze the inference time using the CPU and GPU. We report the average inference time (in seconds) of the AdaSemSeg using NVIDIA TITAN V $12$ GB GPU and Intel Xeon CPU with $128$ GB RAM in Table \ref{tab7:AdaSemSeg_Few_shot_GPU_CPU} for all the datasets studied in this work. We compare the performance of the AdaSemSeg with the prototype-based FSSS method, ProtoSemSeg, and the standard UNet-based segmentation method, the Baseline-$2$ model. We use the same computing infrastructure for all the methods.

As expected, the Baseline-$2$ model (a UNet-based segmentation network) takes the minimum amount of time for both GPU and CPU across all the datasets. Both the AdaSemSeg and ProtoSemSeg take more time than the Baseline-$2$ as the working principle of both the FSSS methods are same. In comparison to ProtoSemSeg, the AdaSemSeg takes more time due to Gaussian process regressions in latent spaces. However, the AdaSemSeg significantly outperforms both ProtoSemSeg and Baseline-$2$ (used in the context of transfer learning) across all datasets and under all evaluation metrics as shown in Table \ref{tab3:AdaSemSeg_Few_shot}. Though the AdaSemSeg is relatively slower than the other methods, the average inference time is comparable and the difference is \textit{not in orders of magnitude}. Considering the trade-off between accuracy and computational time, this analysis justifies the use of AdaSemSeg in real life.

\begin{figure*}[b]
\centering
\includegraphics[width=0.97\textwidth]{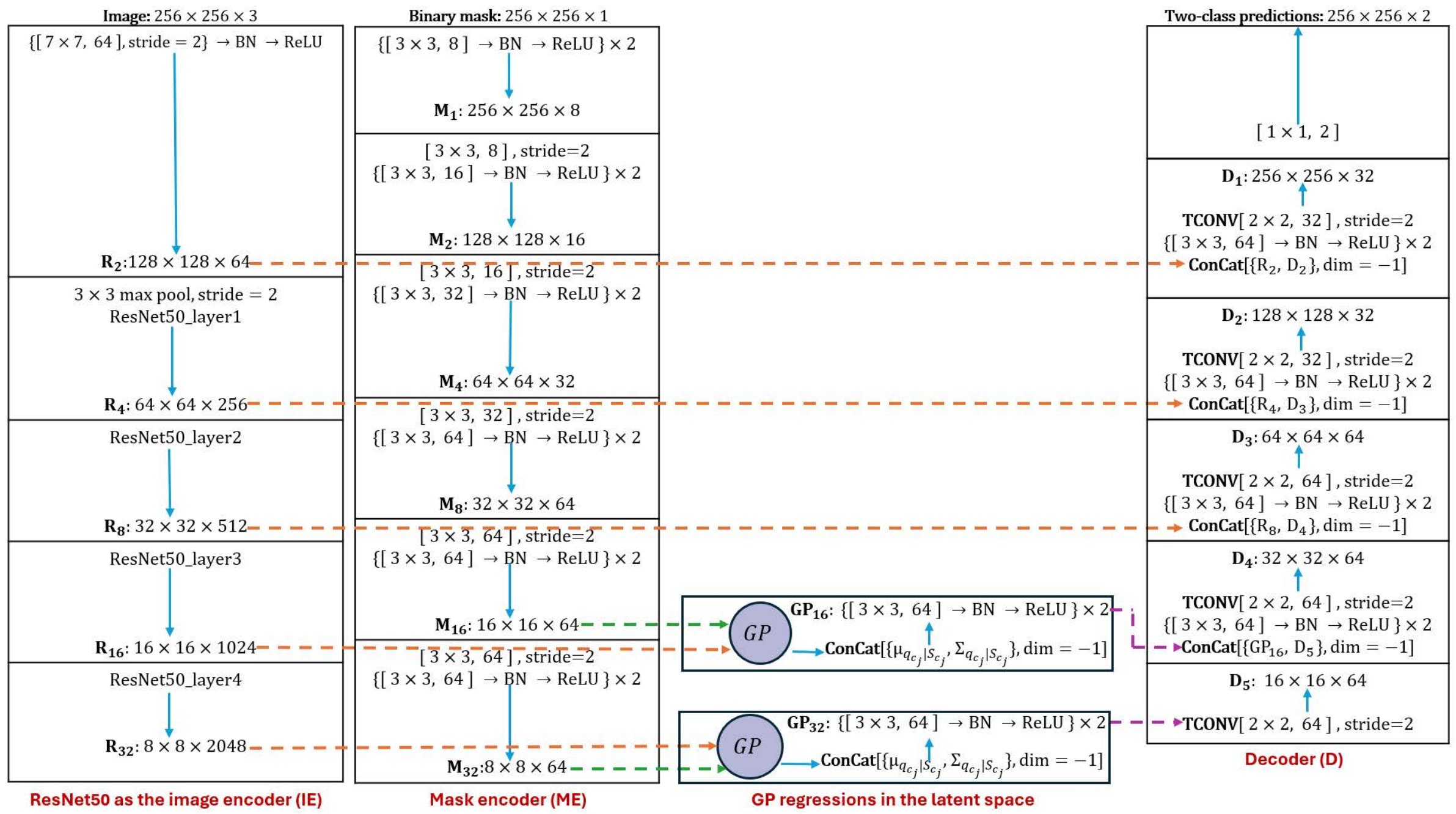}
\caption{The neural network architecture of the AdaSemSeg used to predict the binary mask for a query image.} \label{fig1:AdaSemSeg_Architecture_Details}
\end{figure*}

\section{Conclusion}
This paper introduces a few-shot seismic segmentation (FSSS) method based on Gaussian Process (GP) regression, capable of adapting to varying numbers of facies across different datasets. The proposed AdaSemSeg model is trained on three public seismic datasets\textemdash each with a different number of facies\textemdash without requiring any architectural modifications between datasets. The image encoder is initialized using representations learned through a contrastive learning algorithm applied to the seismic datasets used in this study. This initialization strategy proves more effective than random initialization and enables AdaSemSeg to generalize to unseen target datasets without requiring parameter fine-tuning.

AdaSemSeg achieves performance comparable to baseline models trained directly on the target datasets, particularly on the F3 and Penobscot datasets. Moreover, it outperforms both a competing FSSS method and a segmentation network trained via transfer learning. The ablation studies confirm the robustness and consistency of the proposed method. Comprehensive experiments across all three datasets demonstrate the generalization capability of the AdaSemSeg to new seismic data with $1$ or $5$ annotated examples from the entire 3D volume. These results establish a new state-of-the-art in FSSS for seismic facies interpretation. Despite a slightly higher inference time, the consistent performance of AdaSemSeg supports its practical deployment in real-world seismic interpretation tasks.

{\appendices
\section{Details of the Neural Network Architectures}\label{appsec:NN_arch}
Fig. \ref{fig1:AdaSemSeg_Architecture_Details} shows the detail of the convolutional filters, batch normalization layers, and activation functions used in the image encoder (IE), mask encoder (ME), and decoder (D) of the AdaSemSeg. The IE is the ResNet50 \cite{ResNet_2016}, and the decoder is similar to that used in the U-Net \cite{UNet_2015} with double convolution on the concatenated data followed by transpose convolution. The GP regression takes as input the deep-encoded image features and encoded mask features having the same spatial resolutions. The GP regression is used in two latent layers, i.e., at the bottleneck and the layer above it, whose predictions are fed to the decoder. In addition, the decoder uses shallow encoded image features to predict the binary mask for a query image using support examples.

The Baseline-$2$ and ProtoSemSeg use the IE and D used in the AdaSemSeg with skip connections at multiple layers of encoding as shown in Fig. \ref{fig2:Baseline_Segmentation_ResNet_UNet}. Essentially, this is a U-Net with the ResNet as the encoder, and we call this the ResNet-UNet model. In the case of the ProtoSemSeg \cite{facies_protoseg_2023}, the ResNet-UNet predicts binary mask $C=2$, similar to the AdaSemSeg, and the total loss is accumulated across all the classes in a dataset. However, depending on the dataset, the ResNet-UNet is fixed to $C={6, 7}$ classes for the Baseline-$2$.

\begin{figure*}[b]
\centering
\includegraphics[width=0.97\textwidth]{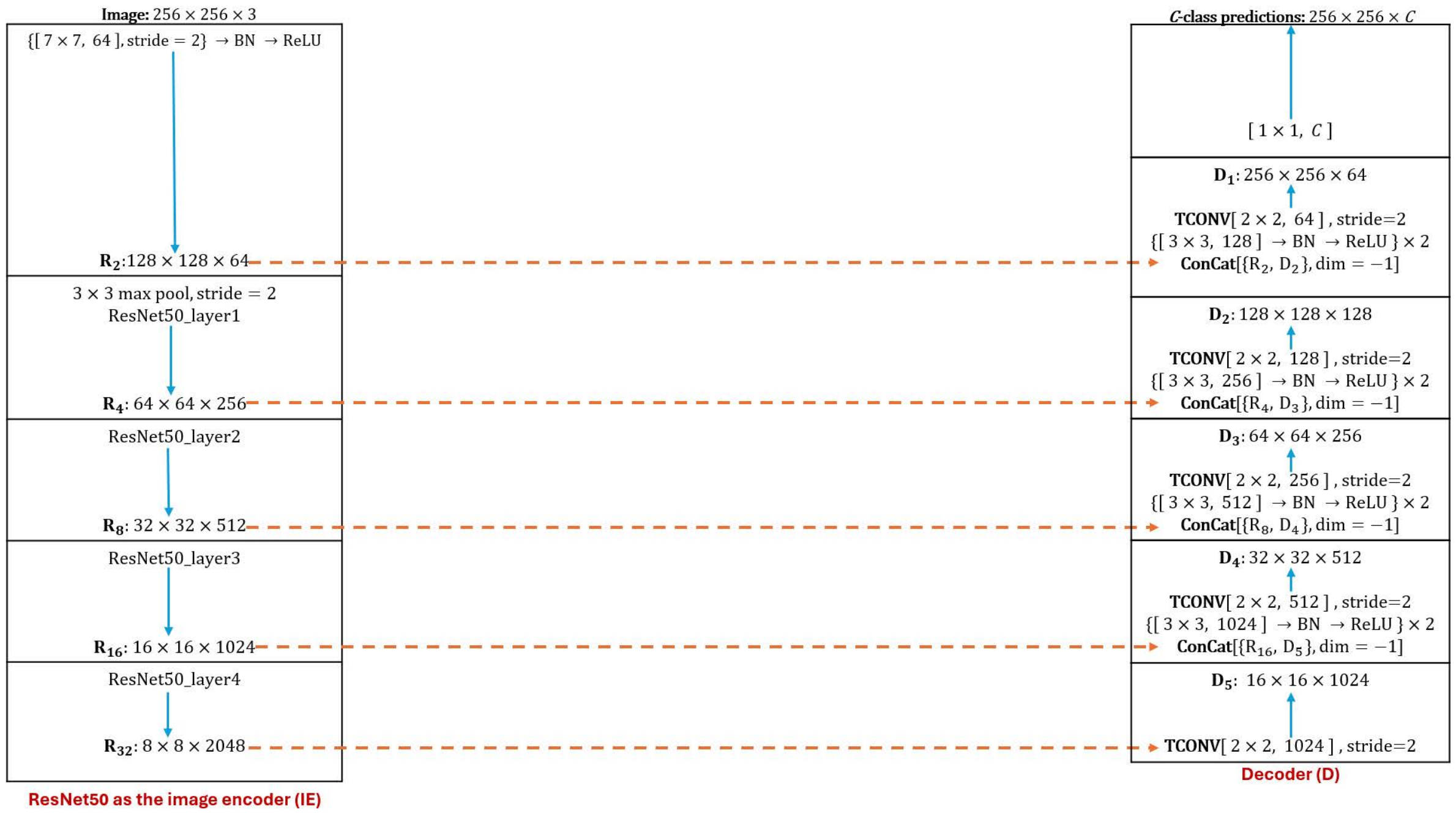}
\caption{The neural network architecture used in Baseline-$2$ and ProtoSemSeg used to predict the multi-class ($C$) mask.} \label{fig2:Baseline_Segmentation_ResNet_UNet}
\end{figure*}

\section{Experiments} \label{appsec:Train_spec}
\subsection{Training the SimCLR} \label{appsec:SimCLRTrain}
The image encoder is trained using SimCLR\cite{SimCLR_2020} on a total of $35648$ patches extracted from the three datasets. Following augmentations used for producing the positive and negative samples: rotation ($[-20^{\circ}, 20^{\circ} ]$), horizontal flip, Gaussian blur ($[0.1, 2.0]$), Gaussian noise ($[1e-4, 5e-2]$), random crop with resize, brightness ($[0.5, 1.5]$), and contrast ($[0.0, 2.0]$). The temperature parameter of the SimCLR is set as $\tau=0.07$. The batch size used in this work is $32$, and we did not observe any performance benefits from a bigger batch size. We used the Adam optimizer with a learning rate of $lr=3e-04$ and weight decay of $1e-04$, and the model was trained for $10$ epochs. The SimCLR achieves an accuracy of $93.75\%$ and $98.44\%$ under top-$1\%$ and top-$5\%$ metric, respectively.

\subsection{Training the AdaSemSeg}
 The GP regression used in the AdaSemSeg uses the configuration used in the DGPNet \cite{dgpnet_2022}. The AdaSemSeg is trained on patches extracted from three datasets. We use $8100$, $7,664$, and $8,206$ patches extracted from the Penobscot, F3, and Parihaka datasets to train the AdaSemSeg under different settings discussed in the paper. We follow the leave-one-out policy to train the AdaSemSeg in a few-shot setup. For example, the source dataset used in the meta-training is built using the Penobscot and F3 datasets, and the trained model is evaluated on the unseen Parihaka dataset. The data augmentations used in this experiment are the RandomRotate ($[-20^{\circ}, 20^{\circ} ]$), RandomHorizontalFlip, GaussianBlur ($[0.1, 2.0]$), and GaussNoise ($[1e-4, 5e-2]$). The image encoder is initialized with the SimCLR-trained statistics. The AdamW optimizer with the learning rate, $lr=5e-05$, and weight decay of $1e-03$ is used. The learning rate is reduced to $0.25$ of the existing rate when the validation loss does not improve for $5$ epochs.

\subsection{Training the Baselines and Competing methods}
The baseline methods used in this work, Baseline-$1$ and Baseline-$2$, and the competing methods, ProtoSemSeg \cite{facies_protoseg_2023} and transfer learning, use the same amount of training data as the AdaSemSeg, under different evaluation scenarios. For the different comparisons studied in this work, all the methods are trained for the same number of epochs with the same configuration for the optimizer and learning rate scheduler. For the ProtoSemSeg, we refer to the settings reported in \cite{facies_protoseg_2023} for all the experimental evaluations.

\begin{table*}[b]
\centering
\caption{Class-wise IoU and $F_1$ scores of the AdaSemSeg to evaluate its sensitivity to under-represented classes across multiple datasets. We present the class distribution (under the Class-wise distribution metric) and highlight under-represented classes\textemdash defined as those comprising $\leq 5\%$ of the total samples\textemdash in \textcolor{blue}{blue}. The corresponding IoU and $F_1$ scores are emphasized in \textcolor{red}{red} and \textcolor{teal}{teal} color, respectively. NA under several class indices for different datasets indicates the absence of the corresponding class indices in the ground truth annotations. The performance of the best method is highlighted in \textbf{bold}.}\label{tab8:AdaSemSeg_sensitivity}
\resizebox{\textwidth}{!}
{
\begin{tabular}{| c | c | c | c c c c c c c c c c c c c c c | c |}
\hline
\multirow{2}{*}{\parbox{1.5cm}{\centering Target Dataset}} & \multirow{2}{*}{Metric} & \multirow{2}{*}{Method} && \multicolumn{13}{c}{Class-wise metric scores} &&\multirow{2}{*}{\parbox{2.0cm}{\centering Weighted Score (FwIoU / Fw$F_1$)}}\\ \cline{5-17}
&&&&1 && 2 && 3 && 4 && 5 && 6 && 7&&\\ \hline
\multirow{5}{*}{\parbox{1.5cm}{\centering Parihaka inline}} &{\parbox{1.5cm}{\centering Class-wise distribution}} 
&&&0.23 &&0.44 &&\textcolor{blue}{0.02} &&0.25 &&0.0 &&\textcolor{blue}{0.05} &&NA &&\\ \cline{2-19}
&\multirow{2}{*}{\parbox{1.5cm}{\centering IoU}}
&AdaSemSeg &&0.85 &&0.73 &&\textcolor{red}{$\mathbf{0.28}$} &&0.77 &&NA &&\textcolor{red}{$\mathbf{0.47}$} &&NA &&0.74\\
&&ProtoSemSeg &&0.31 &&0.33 &&\textcolor{red}{0.03} &&0.55 &&NA &&\textcolor{red}{0.20} &&NA &&0.36\\ \cline{2-19}
&\multirow{2}{*}{\parbox{1.5cm}{\centering $F_1$}}
&AdaSemSeg &&0.92 &&0.85 &&\textcolor{teal}{$\mathbf{0.44}$} &&0.87 &&NA &&\textcolor{teal}{$\mathbf{0.64}$} &&NA &&0.84\\
&&ProtoSemSeg &&0.47 &&0.5  &&\textcolor{teal}{0.06} &&0.71 &&NA &&\textcolor{teal}{0.34} &&NA &&0.52\\ \hline
\multirow{5}{*}{\parbox{1.5cm}{\centering Parihaka crossline}} &{\parbox{1.5cm}{\centering Class-wise distribution}}
&&&0.1 &&0.38 &&\textcolor{blue}{0.03} &&0.33 &&\textcolor{blue}{0.02} &&0.16 &&NA && \\ \cline{2-19}
&\multirow{2}{*}{\parbox{1.5cm}{\centering IoU}}
&AdaSemSeg &&0.62 &&0.71 &&\textcolor{red}{$\mathbf{0.41}$} &&0.80 &&\textcolor{red}{$\mathbf{0.18}$} &&0.62 &&NA &&0.71\\
&&ProtoSemSeg &&0.29 &&0.44 &&\textcolor{red}{0.05} &&0.58 &&\textcolor{red}{0.05} &&0.43 &&NA &&0.46\\ \cline{2-19}
&\multirow{2}{*}{\parbox{1.5cm}{\centering $F_1$}}
&AdaSemSeg &&0.77 &&0.83 &&\textcolor{teal}{$\mathbf{0.58}$} &&0.89 &&\textcolor{teal}{$\mathbf{0.30}$} &&0.77 &&NA &&0.83\\
&&ProtoSemSeg &&0.44  &&0.61  &&\textcolor{teal}{0.09}  &&0.73  &&\textcolor{teal}{0.09}  &&0.61 &&NA &&0.62\\ \hline
\multirow{5}{*}{\parbox{1.5cm}{\centering Penobscot inline}} &{\parbox{1.5cm}{\centering Class-wise distribution}}
&&&0.10 &&0.12 &&\textcolor{blue}{0.03} &&\textcolor{blue}{0.03} &&0.28 &&0.16 &&0.29 && \\ \cline{2-19}
&\multirow{2}{*}{\parbox{1.5cm}{\centering IoU}}
&AdaSemSeg &&0.93 &&0.87 &&\textcolor{red}{$\mathbf{0.75}$} &&\textcolor{red}{$\mathbf{0.67}$} &&0.88 &&0.76 &&0.85 &&0.85\\
&&ProtoSemSeg &&0.4 &&0.37 &&\textcolor{red}{0.0} &&\textcolor{red}{0.04} &&0.37 &&0.52 &&0.50 &&0.42\\ \cline{2-19}
&\multirow{2}{*}{\parbox{1.5cm}{\centering $F_1$}}
&AdaSemSeg &&0.97 &&0.93 &&\textcolor{teal}{$\mathbf{0.86}$} &&\textcolor{teal}{$\mathbf{0.81}$} &&0.94 &&0.87 &&0.92 &&0.93\\
&&ProtoSemSeg &&0.57 &&0.54 &&\textcolor{teal}{0.01} &&\textcolor{teal}{0.07} &&0.54 &&0.68 &&0.66 &&0.58\\ \hline
\multirow{5}{*}{\parbox{1.5cm}{\centering Penobscot crossline}} &{\parbox{1.5cm}{\centering Class-wise distribution}}
&&&0.10 &&0.12 &&\textcolor{blue}{0.02} &&\textcolor{blue}{0.03} &&0.27 &&0.15 &&0.30 && \\ \cline{2-19}
&\multirow{2}{*}{\parbox{1.5cm}{\centering IoU}}
&AdaSemSeg &&0.89 &&0.75 &&\textcolor{red}{$\mathbf{0.57}$} &&\textcolor{red}{$\mathbf{0.65}$} &&0.80 &&0.85 &&0.91 &&0.83\\
&&ProtoSemSeg &&0.33 &&0.26 &&\textcolor{red}{0.0} &&\textcolor{red}{0.06} &&0.39 &&0.42 &&0.48 &&0.38\\ \cline{2-19}
&\multirow{2}{*}{\parbox{1.5cm}{\centering $F_1$}}
&AdaSemSeg &&0.94 &&0.86 &&\textcolor{teal}{$\mathbf{0.72}$} &&\textcolor{teal}{$\mathbf{0.79}$} &&0.89 &&0.92 &&0.95 &&0.90\\
&&ProtoSemSeg &&0.50 &&0.42 &&\textcolor{teal}{0.0}  &&\textcolor{teal}{0.11} &&0.56 &&0.59 &&0.65 &&0.54\\ \hline
\end{tabular}
}
\end{table*}

\section{Additional Results}\label{appsec:Exp_res}
\subsection{Sensitivity to class imbalance across datasets}

Several classes in the datasets analyzed in this study are under-represented, meaning they contain very few annotated examples within the 3D volumes. This leads to a significant class imbalance. To expose the model's sensitivity to the class imbalance issue, we used the intersection over union (IoU) and $F_1$ score, weighted by class frequency as evaluation metrics, denoted as the FwIoU and Fw$F_1$ scores, respectively. While these metrics provide an overall performance summary, they can obscure the model’s behavior on rare classes. To better understand this sensitivity, we report the per-class IoU and $F_1$ scores for two target datasets in Table \ref{tab8:AdaSemSeg_sensitivity}. We also present the class distribution and highlight under-represented classes\textemdash defined as those comprising $\leq 5\%$ of the total samples\textemdash in \textcolor{blue}{blue}. For these classes, we emphasize the IoU and $F_1$ scores in \textcolor{red}{red} and \textcolor{teal}{teal}, respectively. We include the ProtoSemSeg model for comparison.

This analysis focuses on AdaSemSeg trained in a $1-$shot setting on source datasets and evaluated on \textit{unseen} target datasets. We evaluate the AdaSemSeg on the following target datasets: the Parihaka and Penobscot datasets. We must remember that the model parameters of the AdaSemSeg are not finetuned to the target datasets. As shown in Table \ref{tab8:AdaSemSeg_sensitivity}, the AdaSemSeg is \emph{sensitive} to the under-represented classes ($<=5\%$) across datasets and is consistently outperforming the competing method. These findings are further supported by the qualitative predictions shown in Figures \ref{fig6:Parihaka_Predictions} and \ref{fig8:Penobscot_Predictions}, where AdaSemSeg effectively identifies and segments rare classes. This suggests that AdaSemSeg is capable of handling class imbalance robustly. It recognizes and interprets under-represented classes even using a single support example ($1-$shot).

\subsection{Visualizations}
In this section, we discuss the predictions on the Parihaka, and Penobscot datasets by the AdaSemSeg. For the F3 and Penobscot datasets, we use all the support examples (Fig. \ref{fig4:F3_Support_Set} and Fog. Fig. \ref{fig:Penobscot_Support_Set}) for the predictions due to the structural similarity across slices. However, we observe a lot of variation across slices for the Parihaka dataset along both axes, as shown in Fig. \ref{fig5:Parihaka_Variability}. Thus, we resort to the idea of \emph{nearest slice} for the evaluation of test slices, as shown in Fig. \ref{fig7:Parihaka_Support_Set}. Thus, the support set has a single example along each direction, slice index=$\{525\}$ along the inline axis and slice index=$\{657\}$ along the crossline axis. The predictions on the Parihaka dataset are shown in Fig. \ref{fig6:Parihaka_Predictions} that is produced by the AdaSemSeg without training its parameters on the Parihaka dataset and using a \emph{single sample} along each direction (shown in Fig. \ref{fig7:Parihaka_Support_Set}). The AdaSemSeg also does an impressive job in identifying the facies in the Penobscot dataset as shown in Fig. \ref{fig8:Penobscot_Predictions} along the inline and crossline directions using the corresponding support examples in Fig. \ref{fig:Penobscot_Support_Set}.

\begin{figure*}[th]
\centering
\includegraphics[width=0.88\textwidth]{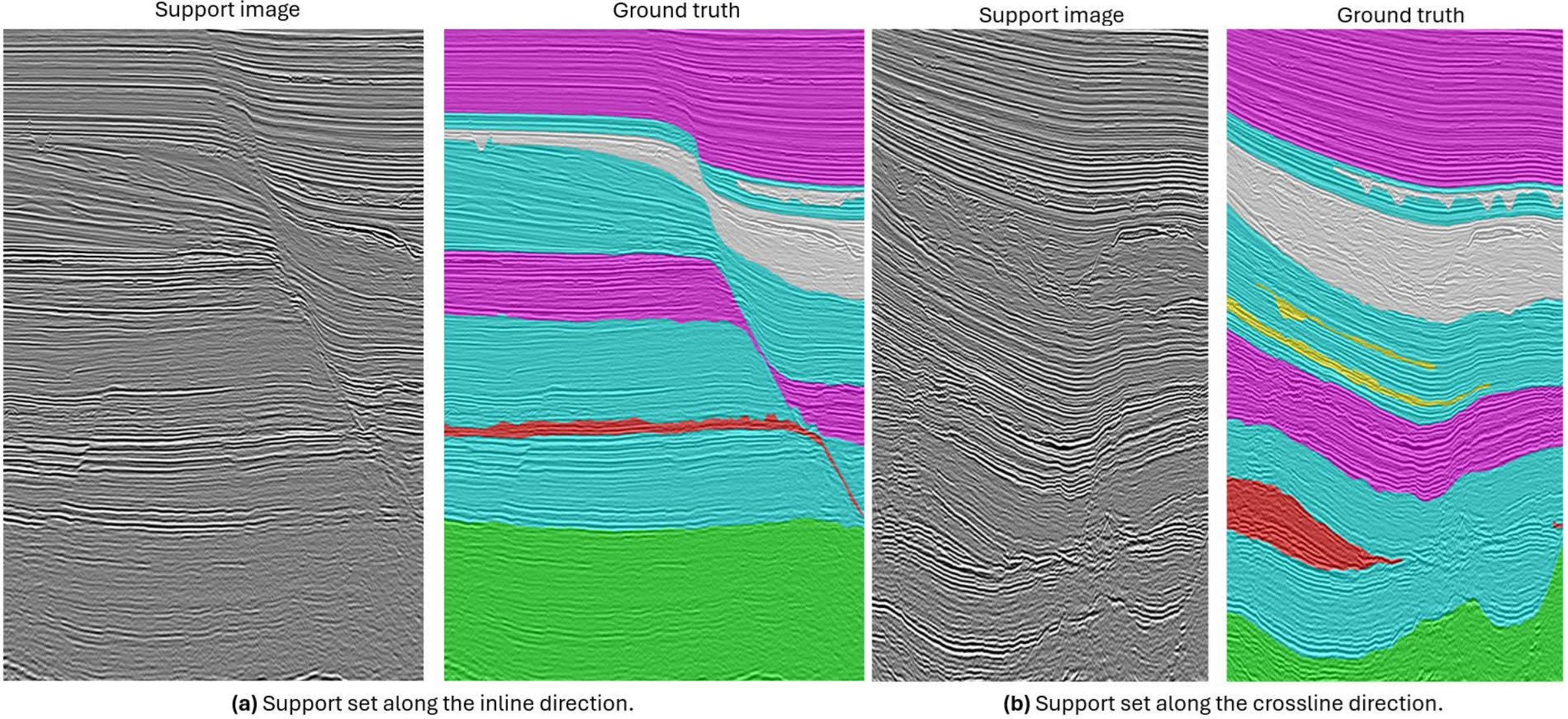}
\caption{The support set for the Parihaka dataset \cite{Parihaka_facies_new_zealand_2020} is the nearest slice to the test data both along the inline (slice index=$\{525\}$) and crossline (slice index=$\{657\}$) directions due to the structural variations along both axes (refer to Fig. \ref{fig5:Parihaka_Variability}). The AdaSemSeg uses the support example along each axis to predict facies, as shown in Fig. \ref{fig6:Parihaka_Predictions}.} \label{fig7:Parihaka_Support_Set}
\end{figure*}

\begin{figure*}[t]
\centering
\includegraphics[width=0.88\textwidth]{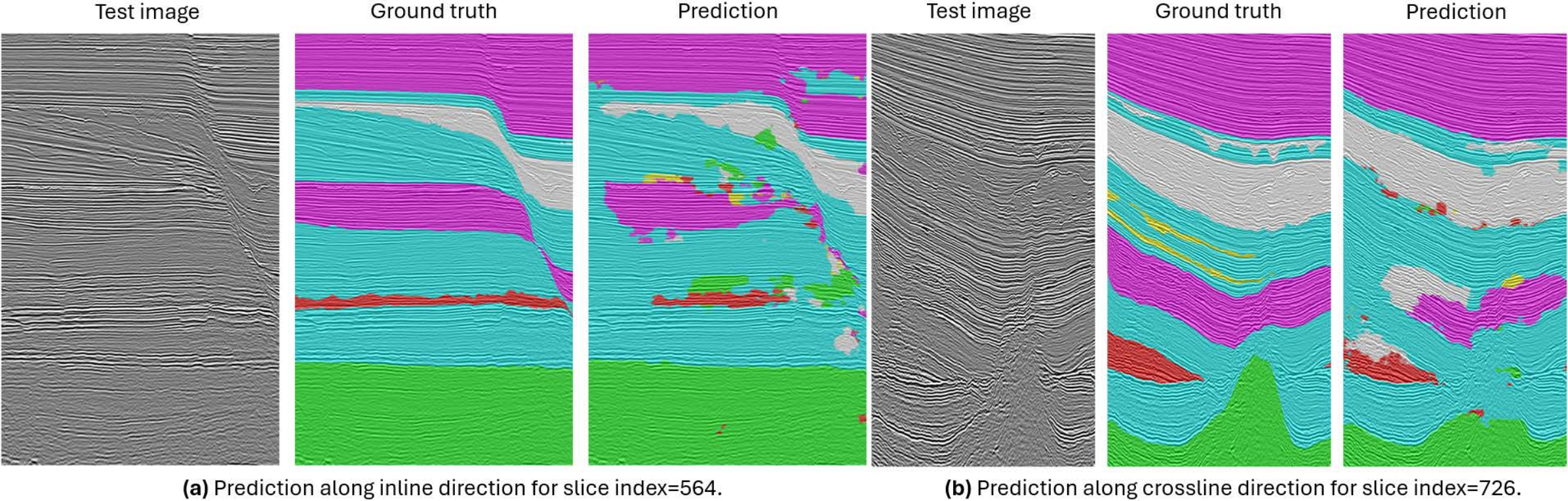}
\caption{Prediction of facies on the test set of the Parihaka dataset \cite{Parihaka_facies_new_zealand_2020} along the inline and crossline directions by the AdaSemSeg trained in the $5-$shot setup on the F3 and Penobscot datasets. The AdaSemSeg uses only $1-$ support examples (as shown in Fig. \ref{fig7:Parihaka_Support_Set}) to predict the facies on the unseen dataset.} \label{fig6:Parihaka_Predictions}
\end{figure*}

\begin{figure*}[th]
\centering
\includegraphics[width=0.88\textwidth]{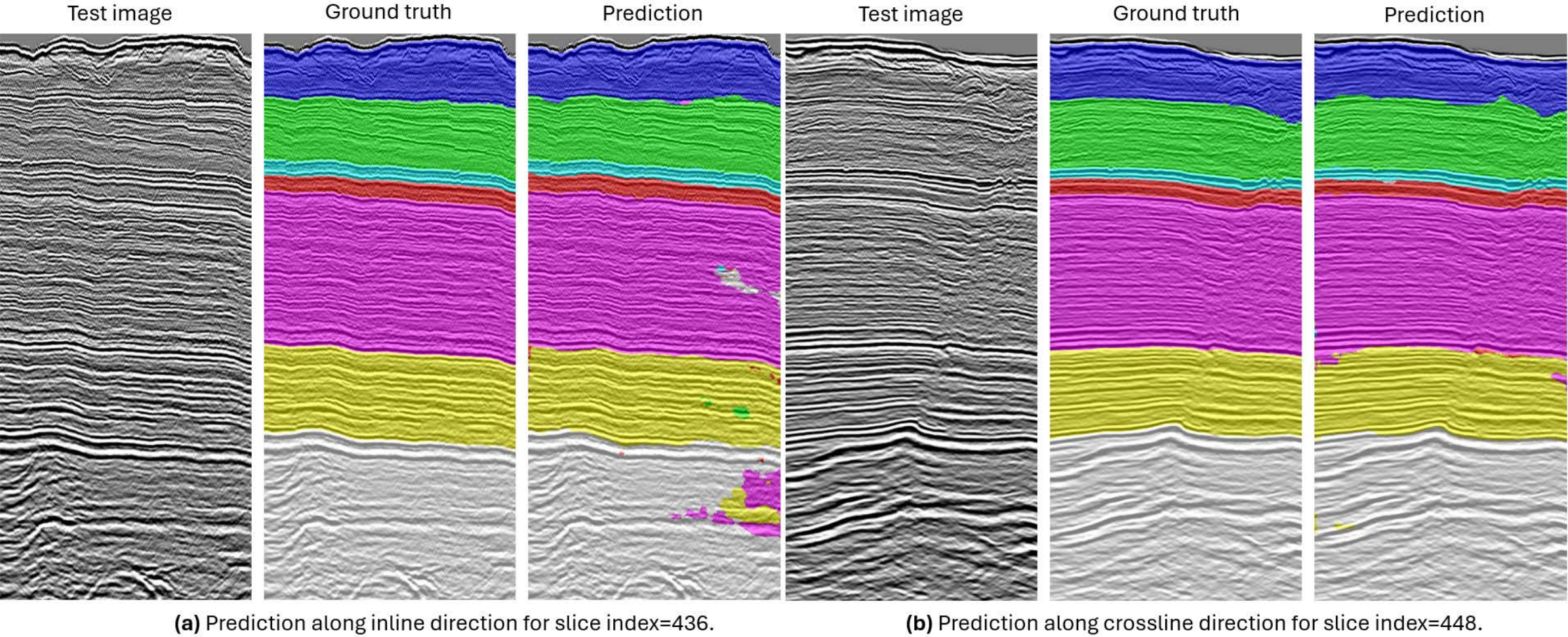}
\caption{Prediction of facies on the test data of the Penobscot dataset \cite{Penobscot_facies_canada_2021} along the inline and crossline axes by the AdaSemSeg trained in the $5-$shot setup on the F3 and Parihaka datasets. The AdaSemSeg uses $5-$ support examples, similar to the F3 dataset, to predict the facies on the unseen dataset.} \label{fig8:Penobscot_Predictions}
\end{figure*}

\begin{figure*}
\centering
\subfloat[The support set of the Penobscot data \cite{Penobscot_facies_canada_2021} along the inline direction.]{{\includegraphics[width=.63\textwidth]{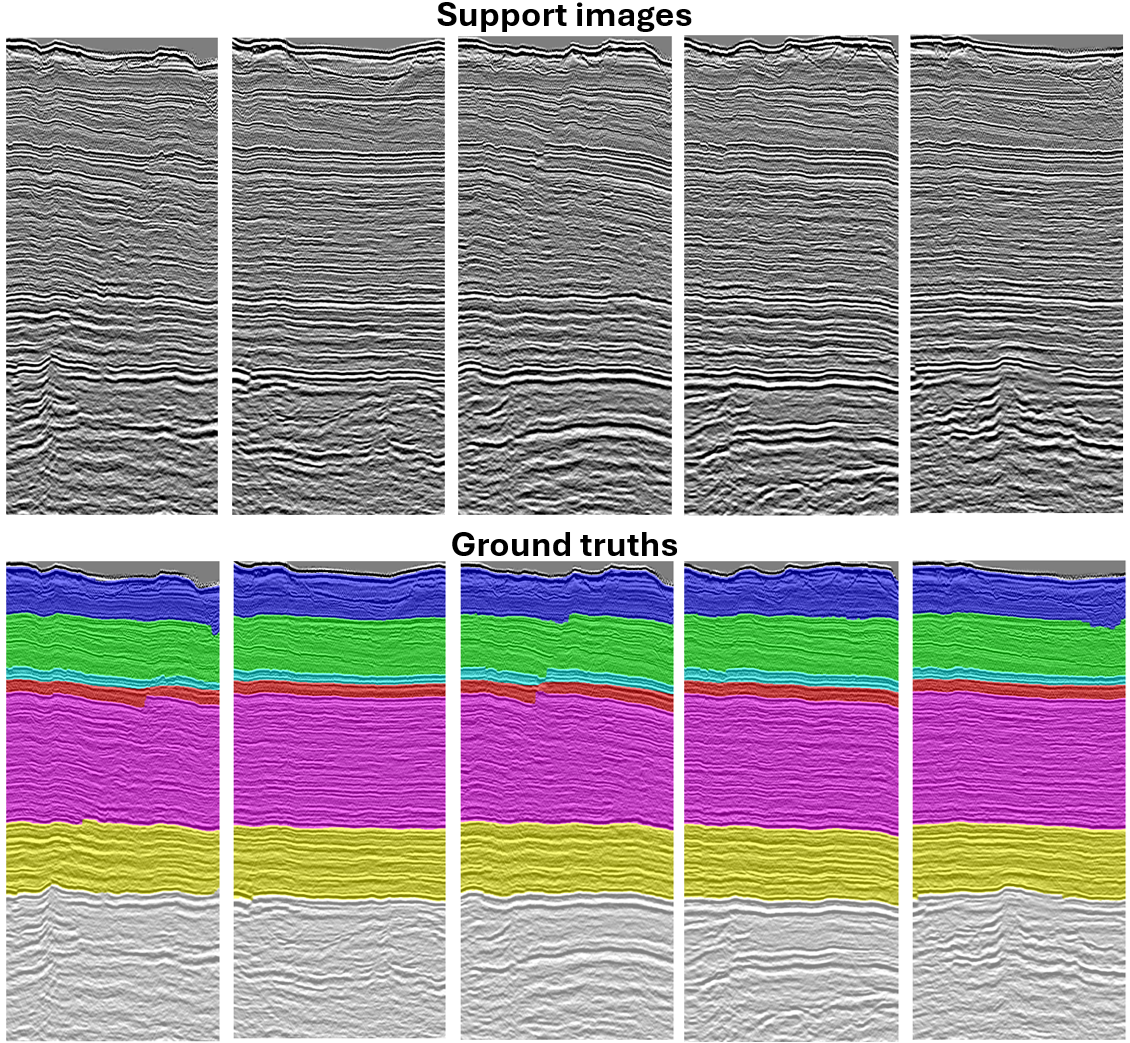} }
\label{fig:Penobscot_Inline_Support_Set}}

\subfloat[The support set of the Penobscot data \cite{Penobscot_facies_canada_2021} along the crossline direction.]{{\includegraphics[width=.63\textwidth]{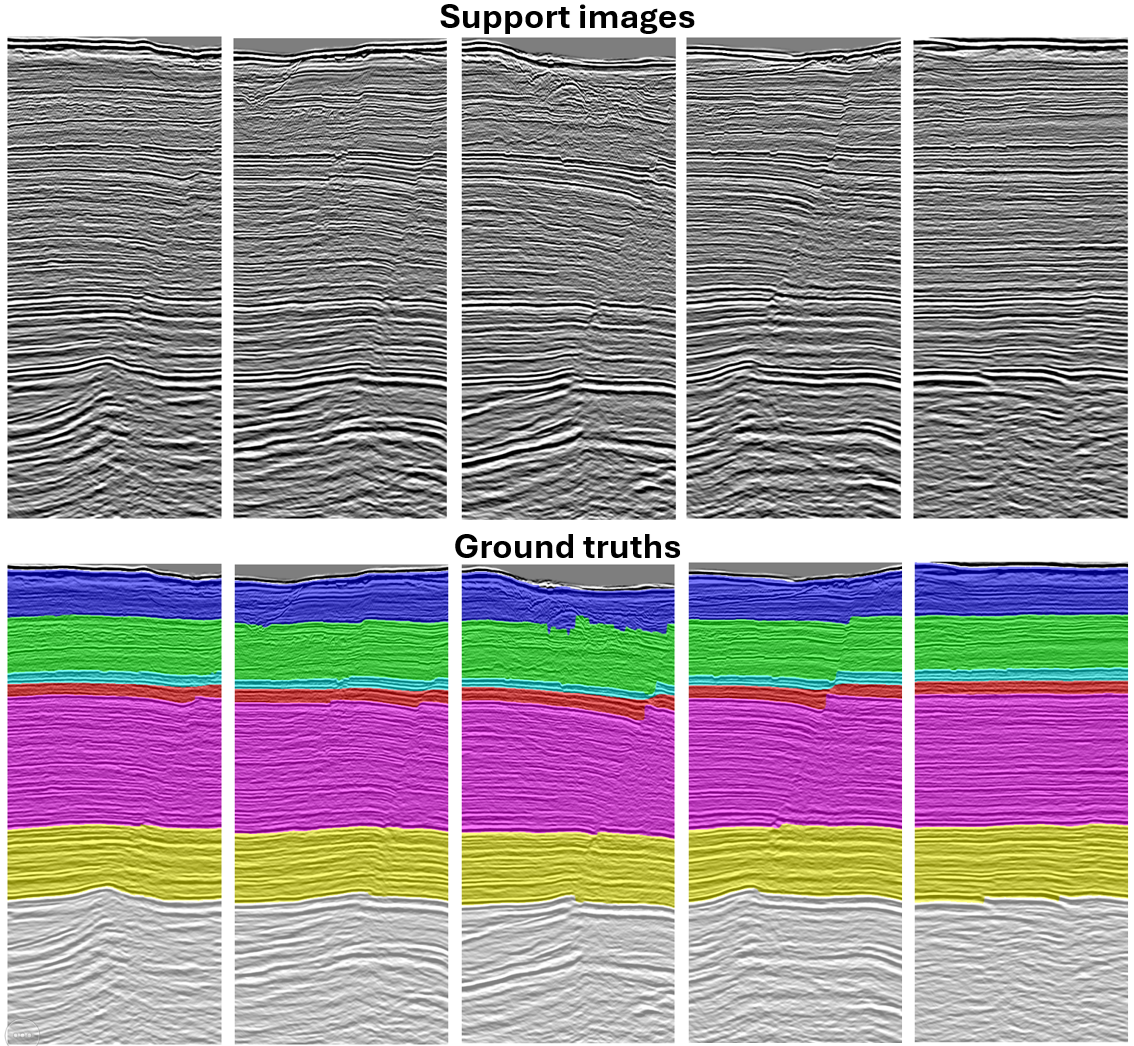} }
\label{fig:Penobscot_Xline_Support_Set}}
\caption{The support set of the Penobscot facies data \cite{Penobscot_facies_canada_2021} that spans the entire volume along the inline and crossline directions. The AdaSemSeg uses the support examples to predict facies on the test slices of thePensobscot dataset shown in Fig. \ref{fig8:Penobscot_Predictions}}
\label{fig:Penobscot_Support_Set}
\end{figure*}


\onecolumn
\twocolumn
\bibliographystyle{IEEEtran}
\bibliography{references}

\end{document}